%% file: main.tex
\documentclass[11pt,logo]{misc/lingqarxiv}

\input{misc/preamble}

\input{misc/command}

\title{\raggedright Foundations of the Scaling Law}

\runningtitle{Foundations of the Scaling Law}

\makeatletter
\newcommand{\twocolumntableofcontents}{%
	\begingroup
	\c@tocdepth=1\relax
	\setlength{\columnsep}{2em}%
	\renewcommand*\l@section[2]{%
		\ifnum \c@tocdepth >\z@
			\addpenalty\@secpenalty
			\addvspace{0.25em}%
			\setlength\@tempdima{1.5em}%
			\begingroup
				\parindent \z@
				\rightskip \@pnumwidth plus 1fil
				\parfillskip -\@pnumwidth
				\leavevmode \bfseries
				\advance\leftskip\@tempdima
				\hskip -\leftskip
				##1\nobreak\hfill\nobreak\hb@xt@\@pnumwidth{\hss ##2}\par
			\endgroup
		\fi
	}%
	\begin{multicols}{2}[
		{\color{GroupBlue}\sffamily\bfseries\fontsize{11}{13}\selectfont\contentsname\endgraf}
		\vspace{-0.5\baselineskip}
	]
		\fontsize{9}{10.2}\selectfont
		\@starttoc{toc}%
	\end{multicols}
	\endgroup
}
\makeatother

\author{%
    Tianyu Zhao\,%
    \href{https://orcid.org/0000-0003-3121-6042}{\faOrcid}%
    \labmark,
    Zhixiang Ren\,%
    \href{https://orcid.org/0000-0002-4104-3790}{\faOrcid}%
    \labmark\textsuperscript{\faEnvelope}\\
    \vspace{1.5ex}
    \labmark LingQiong AI Lab, Shanghai Smart Logic Technology\\
    \faEnvelope~\textbf{Corresponding author:}
    \href{mailto:jason.zhixiang.ren@outlook.com}{%
        \texttt{jason.zhixiang.ren@outlook.com}%
    }
}

\begin{document}

\begin{abstract}
	Empirical scaling laws relate loss to parameters $\PaperN$, training tokens $\PaperD$ and compute $\PaperC$. This technical note explains three mechanisms behind their exponents. Local approximation bounds function error at rate $\PaperN_e^{-\gamma s/d}$, with effective parameters $\PaperN_e$, smoothness $s$, reuse exponent $\gamma$ and dimension $d$; the loss function determines the risk bound. Independent feature coverage gives $\DataExponent=(\SpecExponent-1)/\zeta$ from predictive-weight decay $\SpecExponent$ and occurrence rarity $\zeta$. Separable-loss optimization determines training allocations and deployment crossovers. Nonlinear scoring can produce apparent emergent abilities, while grokking concerns delayed generalization and verification constrains test-time scaling. For regular molecular configurations, the $\mathrm{SE}(3)$ quotient gives $\InternalDim=d-6$ and relative approximation-exponent gain $6/\InternalDim$. Optimized dense-to-equivariant excess-loss ratios scale as $\PaperC^{p_{\mathrm{eq}}-p_{\mathrm{dense}}}$, where $p_j$ are compute-loss exponents and the prefactor depends on loss amplitudes and FLOP costs. Domain examples also cover visual resolution, conformational coverage and correlated-sample precision.
\end{abstract}

\maketitle

\begin{figure}[H]
    \centering
    \includegraphics[width=\linewidth]{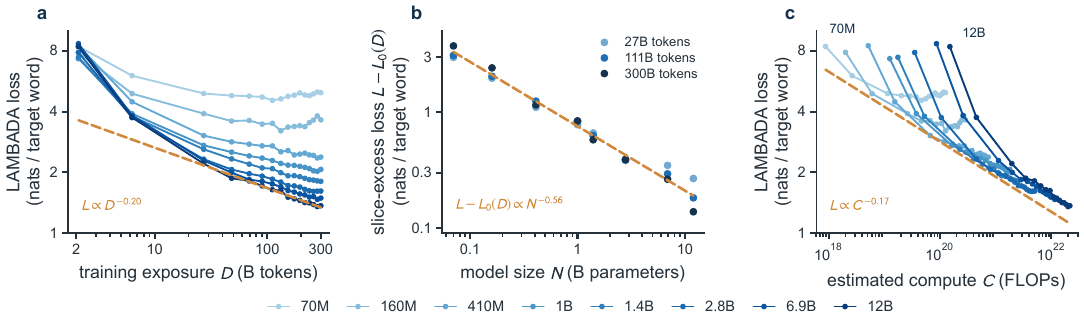}
\caption{\textbf{Pythia scaling across tokens, parameters and compute.} Zero-shot LAMBADA loss $L=\ln\mathrm{PPL}$ (nats per target word) from Pythia~\cite{biderman2023pythia}. \textbf{(a)} Small models plateau as tokens $D$ accumulate; the 12B run improves. \textbf{(b)} Subtracting fitted offsets aligns three token-budget slices of $L-L_0(D)$ around a shared fit. \textbf{(c)} Estimated FLOPs $C\simeq6ND$ compare compute use. Orange dashes trace the 12B late-stage fit (a), shared fit (b) and frontier guide (c).}
    \label{fig:empirical-scaling-overview}
\end{figure}
\twocolumntableofcontents
\clearpage

\input{content/sec1_introduction}
\FloatBarrier
\input{content/sec2_geometric_perspective}
\FloatBarrier
\input{content/sec3_information_theory}
\FloatBarrier
\input{content/sec4_compute_allocation}
\FloatBarrier
\addtocontents{toc}{\protect\columnbreak}
\input{content/sec5_emergence_grokking}
\FloatBarrier
\input{content/sec6_test_time_compute}
\FloatBarrier
\input{content/sec7_beyond_llms}

\FloatBarrier

\FloatBarrier
\phantomsection
\printbibliography[heading=bibintoc,title={References}]

\end{document}

%% file: misc/preamble.tex
\usepackage{graphicx}
\usepackage{float,epstopdf}
\usepackage{bbm}

\usepackage{microtype}
\usepackage{csquotes}

\makeatletter
\def\input@path{{misc/jabbrv/}}
\makeatother
\usepackage[warnundef]{jabbrv}
\DefineJournalAbbreviation{A}{A}
\DefineJournalAbbreviation{X}{X}
\DefineSpuriousJournalWord{The}

\usepackage[
	backend=biber,
	style=numeric-comp,
	sorting=none,
	natbib=true,
	maxnames=6,
	minnames=1,
	maxbibnames=10,
	minbibnames=1,
	eprint=true,
	doi=true,
	url=false,
	isbn=false,
	giveninits=true,
	date=year
]{biblatex}
\AtEveryBibitem{%
	\iffieldundef{eprint}{}{\clearfield{url}\clearfield{urldate}}%
}

\DeclareFieldFormat[article,inproceedings,online]{title}{\mkbibquote{{\rmfamily\fontseries{l}\selectfont #1}}}
\DeclareFieldFormat[book,report,techreport]{title}{\mkbibemph{{\rmfamily\fontseries{l}\selectfont #1}}}
\DeclareFieldFormat[article]{journaltitle}{#1}
\DeclareFieldFormat[inproceedings]{booktitle}{\mkbibemph{#1}}

\DeclareFieldFormat[article]{volume}{\mkbibbold{#1}}
\DeclareFieldFormat[article]{pages}{#1}
\DeclareFieldFormat[article]{eid}{#1}
\renewbibmacro*{volume+number+eid}{%
	\printfield{volume}%
	\iffieldundef{number}
	{}
	{\setunit{\addcomma\space}\printfield{number}}%
	\setunit{\addcomma\space}%
	\printfield{eid}}

\DeclareFieldFormat{doi}{%
	\ifhyperref
	{\href{https://doi.org/#1}{\mkbibacro{DOI}\addcolon\space\nolinkurl{#1}}}
	{\mkbibacro{DOI}\addcolon\space\nolinkurl{#1}}}
\DeclareFieldFormat{eprint:arxiv}{%
	\ifhyperref
	{\href{https://arxiv.org/abs/#1}{%
			arXiv\addcolon\space\nolinkurl{#1}%
			\iffieldundef{eprintclass}
			{}
			{\addspace\texttt{\mkbibbrackets{\thefield{eprintclass}}}}}}
	{arXiv\addcolon\space\nolinkurl{#1}%
		\iffieldundef{eprintclass}
		{}
		{\addspace\texttt{\mkbibbrackets{\thefield{eprintclass}}}}}}

\DeclareSourcemap{
	\maps[datatype=bibtex]{
		\map{
			\pertype{inproceedings}
			\step[fieldsource=booktitle, match=\regexp{.*International\s+Conference\s+on\s+Machine\s+Learning.*}, replace={ICML}]
			\step[fieldsource=booktitle, match=\regexp{.*International\s+Conference\s+on\s+Learning\s+Representations.*}, replace={ICLR}]
			\step[fieldsource=booktitle, match=\regexp{.*Advances\s+in\s+Neural\s+Information\s+Processing\s+Systems.*}, replace={NeurIPS}]
			\step[fieldsource=booktitle, match=\regexp{.*Conference\s+on\s+Computer\s+Vision\s+and\s+Pattern\s+Recognition.*}, replace={CVPR}]
			\step[fieldsource=booktitle, match=\regexp{.*Meeting\s+of\s+the\s+Association\s+for\s+Computational\s+Linguistics.*}, replace={ACL}]
			\step[fieldsource=booktitle, match=\regexp{.*Empirical\s+Methods\s+in\s+Natural\s+Language\s+Processing.*}, replace={EMNLP}]
			\step[fieldsource=booktitle, match=\regexp{.*AAAI\s+Conference\s+on\s+Artificial\s+Intelligence.*}, replace={AAAI}]
			\step[fieldsource=booktitle, match=\regexp{.*Knowledge\s+Discovery\s+and\s+Data\s+Mining.*}, replace={KDD}]
		}
	}
}

\usepackage{subcaption}
\usepackage{booktabs}

\usepackage{amsmath}
\usepackage{amssymb}
\usepackage{mathtools}
\usepackage{amsthm}
\usepackage{dsfont}
\usepackage{multicol}
\usepackage{makecell}
\usepackage{multirow}
\usepackage{amsfonts}
\usepackage{mathrsfs}
\usepackage[amssymb, thickqspace]{SIunits}
\usepackage{enumitem}
\usepackage{pgfplotstable}
\usepackage{lipsum}		%

\usepackage{microtype}
\usepackage{graphicx}
\usepackage{booktabs} %
\usepackage[table]{xcolor}
\usepackage{arydshln}
\usepackage[normalem]{ulem} %

\usepackage{cases}
\usepackage{wrapfig}

\usepackage{url}

\usepackage{thmtools}
\usepackage{thm-restate}
\usepackage{tabu}

\definecolor{huskypurple}{HTML}{4B2E83}

\usepackage{titletoc}

\usepackage{listings}
\lstdefinestyle{promptstyle}{
	basicstyle=\ttfamily\footnotesize,
	breaklines=true,
	breakautoindent=false,
	breakindent=0pt,
	postbreak=\mbox{\textcolor{gray}{$\hookrightarrow$}\space},
	columns=fullflexible,
	keepspaces=true,
	frame=single,
	framesep=5pt,
	xleftmargin=6pt,
	xrightmargin=6pt,
	aboveskip=8pt,
	belowskip=8pt,
	showstringspaces=false,
}

\usepackage[nameinlink]{cleveref}
\crefname{section}{Sec.}{Secs.}
\Crefname{section}{Sec.}{Secs.}
\crefname{subsection}{Sec.}{Secs.}
\Crefname{subsection}{Sec.}{Secs.}
\crefname{subsubsection}{Sec.}{Secs.}
\Crefname{subsubsection}{Sec.}{Secs.}
\crefname{figure}{Fig.}{Figs.}
\Crefname{figure}{Fig.}{Figs.}
\crefname{table}{Tab.}{Tabs.}
\Crefname{table}{Tab.}{Tabs.}
\crefname{equation}{Eq.}{Eqs.}
\Crefname{equation}{Eq.}{Eqs.}
\crefname{appendix}{App.}{Apps.}
\Crefname{appendix}{App.}{Apps.}

\DeclareCiteCommand{\refcite}
  {\ifnumgreater{\value{citetotal}}{1}{Refs.}{Ref.}%
   \addnbspace\bibopenbracket
   \usebibmacro{cite:init}\usebibmacro{prenote}}
  {\usebibmacro{citeindex}\usebibmacro{cite:comp}}
  {}
  {\usebibmacro{cite:dump}\usebibmacro{postnote}\bibclosebracket}

%% file: misc/command.tex
\makeatletter
\def\munderbar#1{\underline{\sbox\tw@{$#1$}\dp\tw@\z@\box\tw@}}
\makeatother

\AddToHook{cmd/appendix/before}{%
	\setcounter{axiom}{0}%
}

\newcommand{\be}{\begin{equation}}
		\newcommand{\ee}{\end{equation}}
\newcommand{\bea}{\begin{equation*}\begin{aligned}}
			\newcommand{\eea}{\end{aligned}\end{equation*}}

\newcommand{\ApproxError}{\eta}
\newcommand{\ClusterCorr}{\varrho_j}
\newcommand{\CovOrder}{s_c}
\newcommand{\DataExponent}{\alpha_D}
\newcommand{\InferFactor}{\rho}
\newcommand{\InferenceCostSaving}{\Delta c_s}
\newcommand{\InternalDim}{d_{\mathrm{int}}}
\newcommand{\OptimalComputeExponent}{\alpha_C}
\newcommand{\PaperC}{C}
\newcommand{\PaperD}{D}
\newcommand{\PaperN}{N}
\newcommand{\ParamExponent}{\alpha_N}
\newcommand{\PartitionFactor}{\chi}
\newcommand{\PrecisionCount}{n_p}
\newcommand{\RigidDim}{d_G}
\newcommand{\SpecExponent}{\beta}
\newcommand{\TargetCoefficient}{a_r}
\newcommand{\TrainFactor}{\kappa}
\newcommand{\TrainingCostDifference}{\Delta C_t}
\newcommand{\UnseenMass}{U}
\newcommand{\VisibilityCutoff}{r_v}
\newcommand{\ChinchillaTokensPerParam}{20}
\newcommand{\GPTThreeParamsBillion}{175}
\newcommand{\GPTThreeTokensBillion}{300}
\newcommand{\GPTThreeTokensPerParam}{1.7}
\newcommand{\KaplanDataExponent}{0.27}
\newcommand{\KaplanParamExponent}{0.73}
\newcommand{\LlamaSmallParamsBillion}{8}
\newcommand{\LlamaTokensPerParam}{1{,}875}
\newcommand{\LlamaTokensTrillion}{15}

%% file: content/sec1_introduction.tex
\section{Introduction: The Empirical Scaling Frontier}
\label{sec:introduction}

Scaling laws quantify the reduction in prediction loss obtained by increasing model size, training data or computation. Under a fixed budget, their slopes and resource costs determine the preferred allocation. Language-model studies estimate these curves by varying parameters, processed training tokens and compute within controlled model families~\cite{kaplan2020scaling,hoffmann2022chinchilla}. \cref{fig:empirical-scaling-overview} shows these dependencies using Pythia's public checkpoint evaluations on a common prediction task~\cite{biderman2023pythia}.

Representation complexity and feature rarity determine different scaling rates. A larger model resolves finer input structure, with the required parameter count increasing with intrinsic dimension and decreasing with target smoothness. Additional data reduces error by revealing features whose contribution depends on both occurrence frequency and predictive importance. Expressing these errors in task-loss units allows their reductions to be balanced against training and inference cost.

Let $\PaperN$ denote active parameter count, $\PaperD$ the number of processed training tokens, and $\PaperC$ training FLOPs. Processed tokens include repeated presentations; the independent-draw models below use sample count instead. Dense training at fixed architectural conventions has approximate cost $\PaperC\approx\TrainFactor\PaperN\PaperD$, where $\TrainFactor$ specifies the FLOP accounting. A one-resource curve holds the other coordinates fixed or optimizes them under a defined budget; execution time additionally depends on the hardware and workload.

\subsection{Power-Law Loss Decay}
\label{subsec:empirical-power-laws}

\begin{figure}[htb!]
 \centering
 \includegraphics[width=0.85\linewidth]{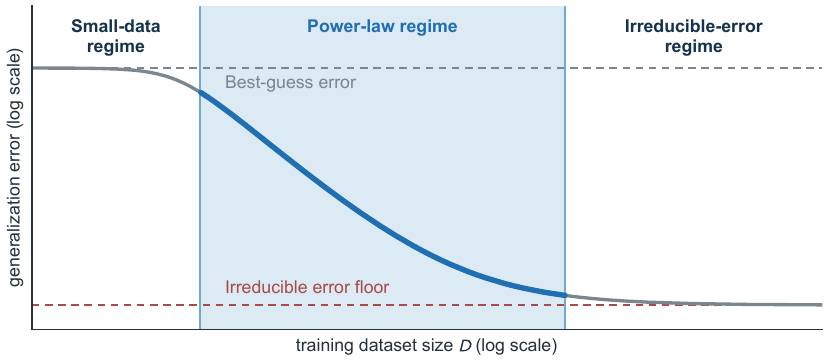}
 \caption{\textbf{Scaling has a finite power-law regime.} As training data $\PaperD$ increase, error passes from a best-guess plateau through power-law decay to the irreducible error floor $L_\infty$, the remaining prediction uncertainty at large data. Redrawn after Hestness et al.~\cite[Fig.~6]{hestness2017scaling}; a single exponent describes the middle interval rather than both plateaus.}
 \label{fig:three-scaling-regimes}
\end{figure}

Constant logarithmic slope characterizes power-law loss decay over a resource interval, as observed across several prediction tasks~\cite{hestness2017scaling,henighan2020autoregressive}. Variance reduction and finer representation resolution provide two mechanisms for such decay~\cite{bahri2021explaining}. Let $X\in\{\PaperN,\PaperD,\PaperC\}$ be the controlled resource, $L_\infty$ the limiting loss floor, and $\epsilon_X(X):=L(X)-L_\infty>0$ the excess loss. A constant logarithmic slope $\alpha_X>0$ over a connected positive resource interval integrates to a power law with amplitude $A_X>0$:
\begin{equation}
 \frac{\mathrm d\log\epsilon_X}{\mathrm d\log X}=-\alpha_X
 \quad\Longleftrightarrow\quad
 L(X)-L_\infty=A_XX^{-\alpha_X}.
 \label{eq:integrated-power-law}
\end{equation}
On log--log axes, the excess loss therefore follows a straight line between the small-data plateau and the error floor shown in \cref{fig:three-scaling-regimes}.

A tenfold increase in $X$ multiplies excess loss by $10^{-\alpha_X}$. The slope therefore measures resource efficiency. Transfer to a new distribution or training protocol can change both the floor and the slope, as transfer studies and broken-slope fits demonstrate~\cite{hernandez2021transfer,caballero2022broken}. An exponent describes the measured model family over its fitted interval.

\subsection{Loss Extrapolation Across Scales}
\label{subsec:extrapolation-boundaries}

Power-law extrapolation predicts larger-scale loss from a fitted floor, amplitude and exponent, using pilot runs within a model family~\cite{rosenfeld2019predictability,openai2023gpt4}. For a proposed scale $\PaperC_f$ beyond the largest pilot $\PaperC_0$, the forecast is
\begin{equation}
 \widehat L(\PaperC_f)=\widehat L_\infty+
 \widehat A_\PaperC\PaperC_f^{-\widehat\alpha_\PaperC}.
 \label{eq:frontier-forecast}
\end{equation}
Held-out pilot scales and repeated seeds test whether the fitted curve predicts beyond the runs used to estimate it. Residual trends reveal a changing slope or floor before the fit is used to allocate a larger budget.

\begin{figure}[htb!]
 \centering
 \includegraphics[width=\linewidth]{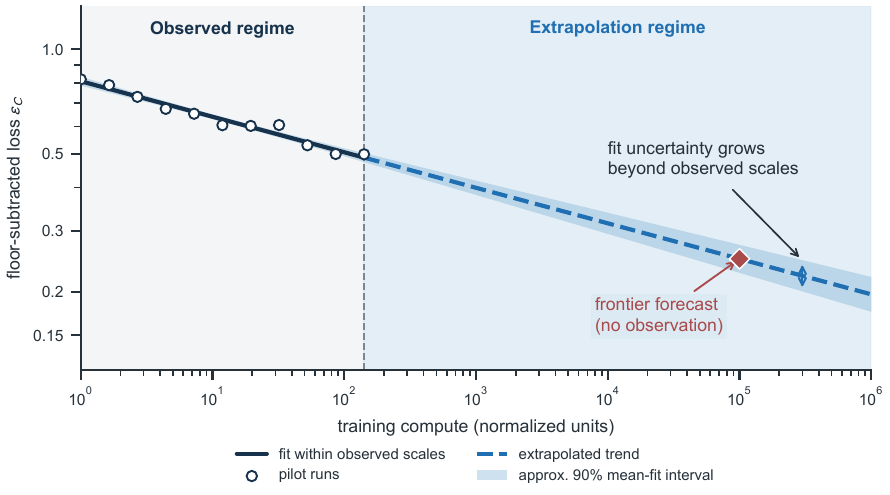}
 \caption{\textbf{Extrapolation magnifies uncertainty in the loss forecast.} Synthetic pilot runs (circles) fit excess loss $\epsilon_C:=L(\PaperC)-L_\infty$ above a fixed floor; solid and dashed lines show the fit and extrapolation to the diamond forecast. The approximate 90\% band for mean log-excess loss widens with extrapolation distance, conditional on the floor and power-law model.}
 \label{fig:scaling-extrapolation}
\end{figure}

Exponent uncertainty amplifies forecast uncertainty with logarithmic distance from the pilot range (\cref{fig:scaling-extrapolation}). Estimating the floor adds uncertainty coupled to the amplitude and slope, while changes in optimization or data distribution can change the curve itself.

\subsection{Resource Cost of Error Reduction}
\label{subsec:bottleneck-reality}

A smaller loss exponent implies a larger multiplicative resource increase for the same fractional error reduction. Differentiating the fitted curve gives $|L'(X)|=\alpha_XA_XX^{-(\alpha_X+1)}$, so marginal gains decline with scale. Inverting the curve gives the resource needed for target excess loss $\epsilon>0$ and the cost of halving that loss:
\[
 X(\epsilon)=\left(\frac{A_X}{\epsilon}\right)^{1/\alpha_X},
 \qquad
 \frac{X(\epsilon/2)}{X(\epsilon)}=2^{1/\alpha_X}.
\]
This multiplier becomes large when $\alpha_X$ is small; comparing resource directions also requires their costs.

\begin{table}[htbp]
 \centering\small
 \caption{How representation, sampling and compute budgets determine error and resource allocation.}
 \label{tab:model-assumptions}
 \begin{tabular}{@{}p{0.19\linewidth}p{0.35\linewidth}p{0.38\linewidth}@{}}
 \toprule
 Model & Assumptions & Prediction or decision\\ \midrule
 Geometry & Compact support, regular cells, local approximation class & Finer cells reduce function error; loss determines the risk bound\\
 Predictive spectrum & Independent draws, aligned feature weights, learning on observation & Error from useful tail features not yet observed\\
 Training allocation & Separable loss, fixed dense FLOP factor & Parameter--data split at fixed training compute\\
 Candidate search & Query-specific proposals and a calibrated selection protocol & Candidate coverage and delivered-answer accuracy\\
 \bottomrule
 \end{tabular}
\end{table}

Finer partitions lower approximation error, new observations reduce error from unseen features, and a fixed compute budget balances the two improvements. The loss function specifies how approximation accuracy enters this allocation calculation.

Throughout this article, analytical curves and schematic maps visualize the models being derived; empirical results are identified by their sources.

%% file: content/sec2_geometric_perspective.tex
\section{Geometry: Resolution, Approximation and Task Risk}
\label{sec:geometric-perspective}

Local polynomial approximation yields a parameter-error bound whose exponent increases with target smoothness and parameter reuse, and decreases with intrinsic dimension. Regular cells resolve the input support, while a polynomial in each cell controls the local remainder. Manifold approximation and resolution-based scaling models formalize this mechanism~\cite{shaham2015manifold,bahri2021explaining}. The loss function then converts the function-error bound into a task-risk bound.

\subsection{Intrinsic Dimension and Covering Complexity}
\label{subsec:intrinsic-manifold}

For manifold-supported data, local covering complexity depends on intrinsic dimension rather than the ambient coordinate count. Let observations lie on a compact support $\mathcal M\subset\mathbb R^m$ with intrinsic dimension $d:=\dim(\mathcal M)$. Under the manifold hypothesis, $d$ can be much smaller than the ambient coordinate count $m$, reducing the number of cells needed for a given resolution. Manifold and compositional regression results establish this reduction for particular support and target classes~\cite{shaham2015manifold,schmidthieber2020regression}; the dimension used here describes the target at the resolution being approximated.

\subsection{Parameter Count and Attainable Resolution}
\label{subsec:parameters-resolution}

The number of attainable local cells determines a model's spatial resolution. For target function $f$ on $\mathcal M$, let $\widehat f_K$ be an approximation fitted over $K$ cells. Its largest discrepancy over the support is the uniform approximation error:
\[
 \ApproxError(f,\widehat f_K):=\|f-\widehat f_K\|_\infty
 =\sup_{x\in\mathcal M}|f(x)-\widehat f_K(x)|.
\]
More cells allow finer spatial resolution. For a fixed architecture family, let $0<\PaperN_e\leq\PaperN$ count the effective parameters used for this resolution and model the attainable cell budget by
\begin{equation}
 K(\PaperN_e)=\PartitionFactor\PaperN_e^\gamma.
 \label{eq:attainable-cell-budget}
\end{equation}
Here $\PartitionFactor>0$ measures partition efficiency and $\gamma>0$ captures parameter reuse. The budget counts cells whose placement and local coefficients the architecture can realize for the target. We evaluate the envelope where $K$ is a positive integer; rounding between these budgets changes only the fixed prefactor.

Depth and composition allow parameters to be reused across local computations~\cite{montufar2014regions,raghu2016expressivity}, while equivariance shares representations across symmetry-related inputs~\cite{cohen2016equivariant}. These mechanisms motivate the cell budget, but realizing a target approximation also requires control of cell geometry and coefficients. Thus $\PartitionFactor$ measures usable partition capacity; the separate coefficient $\TrainFactor$ measures training FLOPs.

\Needspace{10\baselineskip}
\subsection{From Spatial Resolution to Function Error and Task Loss}
\label{subsec:geometric-limit}

On a regular $d$-dimensional partition, smoothness $s$ gives an approximation-error bound proportional to $K^{-s/d}$. The derivation combines cell diameter with the local Taylor remainder. A cell of diameter $h$ occupies a volume proportional to $h^d$, so a regular partition into $K$ cells has resolution proportional to $K^{-1/d}$. Writing $h_K$ for the largest cell diameter in local coordinates gives the covering bound
\begin{equation}
 h_K\leq C_{\mathcal M}K^{-1/d}.
 \label{eq:covering-resolution}
\end{equation}
This construction uses a smooth compact manifold with a finite atlas and cells of comparable size and bounded aspect ratio. The constant $C_{\mathcal M}>0$ accounts for the geometry, chart distortion and partition regularity, independently of $K$.

Target smoothness determines the error remaining within each cell. Smoothness of order $s>0$ means that a local Taylor polynomial of degree $\lceil s\rceil-1$ leaves a remainder bounded by a constant times the cell diameter raised to $s$. For targets in a fixed H\"older ball $\mathcal H_s$, this gives
\begin{equation}
 \ApproxError\leq c_sh_K^s
 \leq c_sC_{\mathcal M}^sK^{-s/d}.
 \label{eq:local-taylor-rate}
\end{equation}
The H\"older ball fixes a common bound on the derivatives and their continuity; its remainder constant $c_s>0$ is independent of $K$. For noninteger $s$, the highest Taylor derivative has H\"older exponent $s-\lceil s\rceil+1$; at integer $s$, the corresponding Lipschitz remainder condition suffices.

\begin{figure}[tbp]
 \centering
 \includegraphics[width=\linewidth]{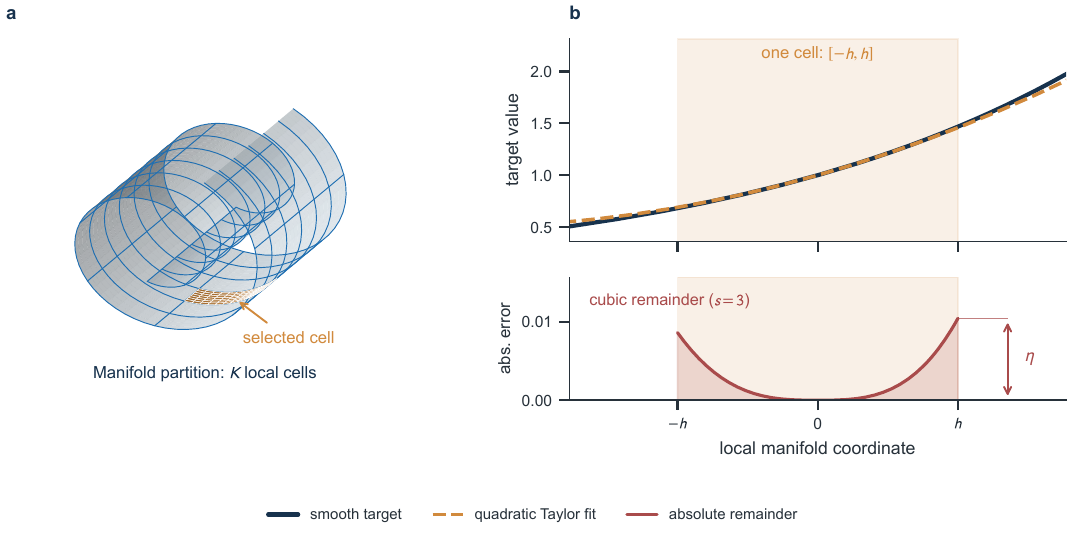}
 \caption{\textbf{Finer cells and smoother targets reduce approximation error.} \textbf{(a)} A $d$-dimensional manifold is covered by $K$ regular cells with size $h\propto K^{-1/d}$. \textbf{(b)} A quadratic Taylor approximation (dashed orange) to a smooth target (navy) leaves a uniform error $\ApproxError$ (maximum red residual) bounded by a constant times $h^3$. More generally, smoothness $s$ gives a bound scaling as $K^{-s/d}$.}
 \label{fig:dimensional-tax}
\end{figure}

The attainable cell budget converts spatial refinement into a parameter-error bound. Substituting \cref{eq:attainable-cell-budget} into \cref{eq:local-taylor-rate} and defining $A_g:=c_sC_{\mathcal M}^s\PartitionFactor^{-s/d}$ gives
\begin{equation}
 \ApproxError(\PaperN_e)\leq A_g\PaperN_e^{-\gamma s/d}.
 \label{eq:parameter-error-bound}
\end{equation}
For a Lipschitz target ($s=1$), linear cell growth ($\gamma=1$) and full parameter use ($\PaperN_e=\PaperN$), this becomes $\ApproxError(\PaperN)\leq A_g\PaperN^{-1/d}$. Higher smoothness improves the local fit; higher dimension requires more cells to reach the same resolution. \cref{fig:dimensional-tax} illustrates the partition and the local remainder.

For the fixed-degree piecewise-polynomial space $\mathcal P_K$ on this partition, the best uniform approximation error, maximized over $\mathcal H_s$, lies between $c_-K^{-s/d}$ and $c_+K^{-s/d}$, with positive constants independent of $K$. A rescaled smooth bump gives the lower bound and Taylor approximation gives the upper bound. This matching rate concerns that local approximation class; a network realizing it inherits the upper bound. Neural approximation results specify their own architecture and target classes~\cite{yarotsky2017error}, and deep or adaptive constructions can have different rates~\cite{yarotsky2018continuous,suzuki2018adaptive}.

The loss determines whether function approximation error enters the risk bound linearly or quadratically. For an input--response pair $(x,y)$ and predictor $g$, define $\mathcal R(g):=\mathbb E[\ell(g(x),y)]$, with finite risks for the predictors being compared. \textit{Lipschitz losses:} if $\ell(\widehat f(x),y)$ is $G$-Lipschitz in its prediction argument over both predictors' output range, then
\begin{equation}
 |\mathcal R(\widehat f)-\mathcal R(f)|\leq G\ApproxError.
 \label{eq:approximation-risk-transfer}
\end{equation}
This transfers an $h^s$ uniform function-error bound to an $h^s$ upper bound on the risk difference, with $G$ fixed.

\textit{Squared loss:} assume $\mathbb E[y^2]<\infty$ and take the approximation target to be the Bayes regression function $f^\star(x)=\mathbb E[y\mid x]$. Conditional expectation is an $L^2$ contraction, so $f^\star$ is square-integrable; a finite uniform error $\ApproxError$ then makes $\widehat f$ square-integrable under the input probability measure. Cauchy--Schwarz makes the cross term integrable, and conditional expectation eliminates it:
\[
 \mathbb E[(y-f^\star)(\widehat f-f^\star)]
 =\mathbb E\!\left[\mathbb E[y-f^\star\mid x]
 (\widehat f-f^\star)\right]=0.
\]
Expanding squared loss therefore yields the finite-risk identity
\begin{equation}
 \mathcal R(\widehat f)-\mathcal R(f^\star)
 =\mathbb E[(\widehat f-f^\star)^2]\leq\ApproxError^2.
 \label{eq:squared-risk-transfer}
\end{equation}
Thus an $h^s$ function-error rate can yield an $h^{2s}$ upper bound for squared risk.

\textit{Categorical cross-entropy:} for finitely many classes, let $p(x)$ and $\widehat p(x)$ be normalized true and predicted conditional probability vectors. Assume a common lower bound $\widehat p_i(x)\geq\delta_0>0$ for every input and class. Writing $\operatorname{KL}$ for Kullback--Leibler divergence, the inequality $\log u\leq u-1$ for $u>0$ gives
\begin{equation}
 \mathcal R_{\mathrm{CE}}(\widehat p)-\mathcal R_{\mathrm{CE}}(p)
 =\mathbb E_x\operatorname{KL}(p(x)\|\widehat p(x))
 \leq\delta_0^{-1}\mathbb E_x\|p(x)-\widehat p(x)\|_2^2.
 \label{eq:cross-entropy-risk-transfer}
\end{equation}
The probability lower bound controls the logarithm near zero, so squared probability-vector error bounds excess cross-entropy. Together, the three branches show why the loss function matters: the same function-error bound can give a linear risk bound or a quadratic one.

\subsection{Comparing Representations at Matched Quality}
\label{subsec:geometric-engineering-implication}

Reducing $d$, increasing $\gamma$, or improving $\PartitionFactor$ lowers the parameter budget needed to reach a given function error. Retrieval, routing and symmetry constraints can achieve such changes by altering the representation and its approximation class. Their resource benefit is the total preprocessing, training and serving cost saved at matched task quality, compared with simply increasing $\PaperN$.

\begin{insight}[Dimension, Smoothness and Resolution Cost]
 Halving the function-error upper bound requires a region-budget multiplier $2^{d/s}$, or an effective-parameter multiplier $2^{d/(\gamma s)}$, in the continuous power-law envelope. Lipschitz losses inherit the function-error exponent, while squared loss at the Bayes target doubles the exponent of the risk upper bound.
\end{insight}

%% file: content/sec3_information_theory.tex
\section{Information Spectra and Data Scaling}
\label{sec:information-theory}

In a feature-coverage model, data-loss decay is determined by the balance between feature rarity and predictive importance. Common features are observed early, leaving rarer features to contribute the remaining error. Covariance and target spectra specify these contributions, while regularization determines which modes are resolved at a given sample size~\cite{bordelon2020spectrum,canatar2021alignment}.

\subsection{Feature Observation and the Visibility Scale}
\label{subsec:quantization-knowledge}

The expected feature count sets the transition between predominantly unseen and predominantly observed ranks. Let $q(r)$ be the occurrence probability of feature rank $r=1,2,\ldots$, and let $\lambda(r)$ be the error contribution that remains until that feature is learned, termed its predictive mass. Occurrence probability determines coverage, while predictive mass determines the loss associated with missing the feature. Covariance eigenvalues model these contributions when the target has power in the corresponding modes~\cite{canatar2021alignment}.

For $\PaperD$ independent draws, the probability of seeing feature $r$ at least once is
\begin{equation}
 P_r(\PaperD)=1-(1-q(r))^{\PaperD}
 \approx1-\exp[-\PaperD q(r)].
 \label{eq:feature-observation-probability}
\end{equation}
For rare features with small $q(r)$, the exponential approximates the exact probability. The transition occurs near an expected count $\PaperD q(r)\approx1$: below it the feature is usually absent, and above it observation becomes likely. The model below treats one observation as sufficient to learn the feature; noisy estimation and correlated samples require a richer learning model.

\subsection{Data Scaling from Unseen Features}
\label{subsec:zipf-feature-spectra}

Under independent sampling and learning on first observation, power-law feature frequencies and error weights yield data-loss exponent $(\SpecExponent-1)/\zeta$~\cite{hutter2021learning}. Take
\begin{equation}
 q(r)=Z^{-1}r^{-\zeta},\qquad
 \lambda(r)=Br^{-\SpecExponent},\qquad
 \zeta>1,\quad\SpecExponent>1,\quad B>0.
 \label{eq:zipf-spectral-model}
\end{equation}
The occurrence exponent $\zeta$ controls rarity, the weight exponent $\SpecExponent$ controls predictive importance, and $B$ sets its amplitude. Here $Z=\sum_{r=1}^\infty r^{-\zeta}$ normalizes occurrence probabilities. Under independent draws, exact learning on first observation and additive feature errors, the expected error from features not yet observed is
\begin{equation}
 \UnseenMass(\PaperD)=\sum_{r=1}^\infty
 Br^{-\SpecExponent}(1-Z^{-1}r^{-\zeta})^{\PaperD}.
 \label{eq:smooth-unseen-mass}
\end{equation}
Each term multiplies a feature's error contribution by its probability of remaining unseen, using \cref{eq:feature-observation-probability}. More detailed learning curves also include noise, repeated estimation and regularization~\cite{spigler2019learning,caponnetto2007rates}.

Poissonizing the draw count with mean $\PaperD$ makes the unseen probability exactly $\exp[-\PaperD q(r)]$. In the continuum approximation to the rank sum, put $t=\PaperD/Z$ and $p=(\SpecExponent-1)/\zeta>0$. The change of variable and its Jacobian are
\[
 u=tr^{-\zeta},\qquad
 r=\left(\frac{t}{u}\right)^{1/\zeta},\qquad
 \mathrm dr=-\frac{t^{1/\zeta}}{\zeta}
 u^{-1-1/\zeta}\,\mathrm du.
\]
Substitution into the rank integral gives
\begin{equation}
 \begin{aligned}
 I(t)&=\int_1^\infty Br^{-\SpecExponent}e^{-tr^{-\zeta}}\,\mathrm dr\\
 &=\frac{B}{\zeta}t^{-p}\int_0^t u^{p-1}e^{-u}\,\mathrm du.
 \end{aligned}
 \label{eq:poissonized-continuum-tail}
\end{equation}
The incomplete Gamma integral converges to $\Gamma(p)$ as $t\to\infty$. Write $\UnseenMass_c(\PaperD):=I(\PaperD/Z)$ for the continuum residual. Substitution retains the full leading coefficient:
\begin{equation}
 \UnseenMass_c(\PaperD)\sim
 \frac{B\Gamma(\DataExponent)}{\zeta}Z^{\DataExponent}\PaperD^{-\DataExponent},
 \qquad\DataExponent=\frac{\SpecExponent-1}{\zeta}.
 \label{eq:data-scaling-spectral}
\end{equation}
Measuring the feature contributions in loss units sets $L(\PaperD)-L_\infty:=\UnseenMass_c(\PaperD)$ in this continuum model. A steeper importance spectrum makes the unseen tail less costly, while a steeper rarity spectrum makes it slower to observe: their balance gives the data exponent. Since $t=\PaperD/Z$, the normalization contributes $Z^{\DataExponent}$ to the leading coefficient.

The omitted tail is controlled by the upper incomplete Gamma function $\Gamma(p,t):=\int_t^\infty u^{p-1}e^{-u}\,\mathrm du$:
\[
 \frac{B}{\zeta}\Gamma(p)t^{-p}-I(t)
 =\frac{B}{\zeta}t^{-p}\Gamma(p,t)
 \sim\frac{B}{\zeta t}e^{-t}
 \qquad(t\to\infty).
\]

\begin{figure}[p]
 \centering
 \includegraphics[width=\linewidth]{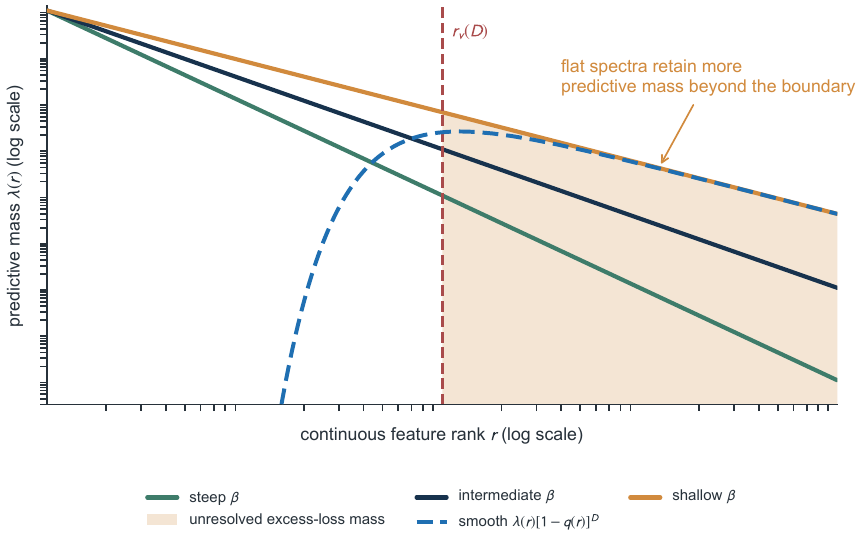}
 \caption{\textbf{Feature visibility converts spectral tails into data scaling.} Rank $r$ carries predictive mass $\lambda(r)\propto r^{-\SpecExponent}$ and occurrence probability $q(r)\propto r^{-\zeta}$; sample count $\PaperD$ sets the boundary $\PaperD q(\VisibilityCutoff)=1$. For the shallow spectrum, hard-tail shading and smooth unseen mass $\lambda(r)[1-q(r)]^{\PaperD}$ (blue dashed) yield the same large-data exponent $(\SpecExponent-1)/\zeta$ in the coverage model.}
 \label{fig:zipf-spectra}
 \vspace{1em}
 \includegraphics[width=\linewidth]{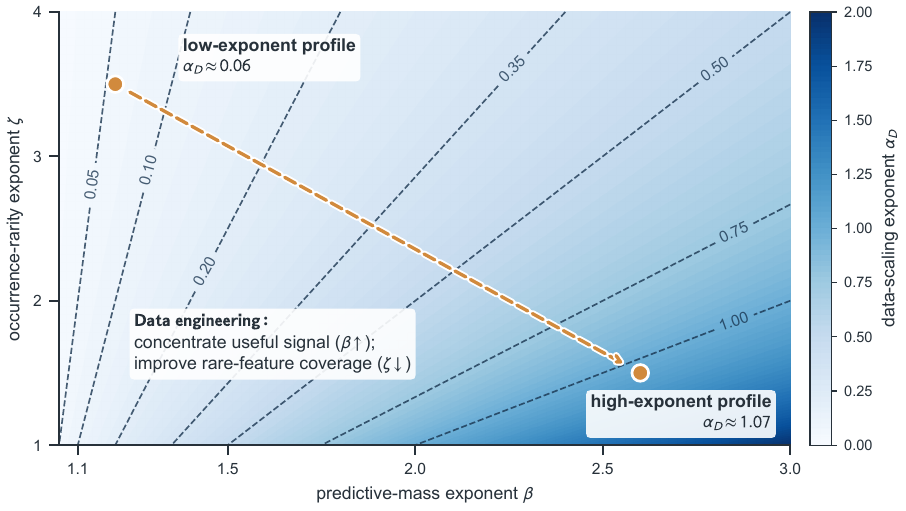}
 \caption{\textbf{Predictive concentration and feature rarity jointly set the data exponent.} Blue shading and labeled contours give $\DataExponent=(\SpecExponent-1)/\zeta$, where $\SpecExponent$ controls predictive-mass decay across feature ranks and $\zeta$ controls occurrence-frequency decay. Orange points and the dashed arrow compare two profiles: concentrating mass toward low ranks (larger $\SpecExponent$) and making tail features less rare (smaller $\zeta$) raise the predicted loss-decay exponent.}
 \label{fig:scaling-phase-space}
\end{figure}

A hard cutoff locates the same transition by setting the expected occurrence count to one:
\[
 \PaperD q(\VisibilityCutoff)=1
 \quad\Longrightarrow\quad
 \VisibilityCutoff=\left(\frac{\PaperD}{Z}\right)^{1/\zeta}
 =t^{1/\zeta}.
\]
Replacing the smooth unseen probability by the indicator $r>\VisibilityCutoff$ produces
\begin{equation}
 \begin{aligned}
 H(t)&=\int_{t^{1/\zeta}}^\infty Br^{-\SpecExponent}\,\mathrm dr
 =\frac{B}{\SpecExponent-1}t^{-p},\\
 \frac{I(t)}{H(t)}&\longrightarrow p\Gamma(p)=\Gamma(1+p).
 \end{aligned}
 \label{eq:smooth-hard-tail-ratio}
\end{equation}
The smooth and hard descriptions share an exponent, with generally different constants. \cref{fig:zipf-spectra} compares the hard-tail shading with the smoothly weighted unseen mass; \cref{fig:scaling-phase-space} maps the common exponent across spectral and frequency slopes.

Finite support changes the late-data regime. With final rank $r_{\max}$ and $1\leq\VisibilityCutoff\leq r_{\max}$, the hard-tail mass is
\[
 \int_{\VisibilityCutoff}^{r_{\max}}Br^{-\SpecExponent}\,\mathrm dr
 =\frac{B}{\SpecExponent-1}
 \left(\VisibilityCutoff^{1-\SpecExponent}-r_{\max}^{1-\SpecExponent}\right).
\]
It becomes zero once the cutoff reaches $r_{\max}$. The smooth unseen mass instead approaches zero continuously; with finitely many positive occurrence probabilities, its asymptotic decay is exponential. For $\SpecExponent\leq1$, the infinite-rank predictive mass diverges, so a finite-support model or a different tail is required.

\subsection{From Dimension to Spectral Decay}
\label{subsec:weyls-law}

For an inverse-Laplacian covariance, the spectral decay exponent is twice the covariance order divided by intrinsic dimension. Weyl's law determines the growth of the underlying spatial frequencies~\cite{weyl1912eigenvalues}. Let $\mathcal M$ be a smooth compact $d$-dimensional Riemannian manifold without boundary, and use the nonnegative self-adjoint Laplace--Beltrami operator $\Delta_{\mathcal M}=-\operatorname{div}_g\nabla_g$ on $L^2(\mathcal M,\mathrm{vol}_g)$, with Sobolev domain $H^2(\mathcal M)$ and eigenvalues $\mu_r\geq0$. A positive shift $a>0$ and covariance order $\CovOrder>d/2$ define the positive trace-class covariance~\cite{borovitskiy2020matern}
\[
 T=(aI+\Delta_{\mathcal M})^{-\CovOrder}.
\]
Let $\omega_d$ be the Euclidean unit-ball volume and define $w_{\mathcal M}:=(2\pi)^2[\omega_d\operatorname{Vol}(\mathcal M)]^{-2/d}$. Substituting Weyl's eigenvalue growth into the covariance gives
\begin{equation}
 \begin{aligned}
 \mu_r&\sim w_{\mathcal M}r^{2/d},\\
 \lambda_r=(a+\mu_r)^{-\CovOrder}
 &\sim w_{\mathcal M}^{-\CovOrder}r^{-2\CovOrder/d}
 \qquad(r\to\infty).
 \end{aligned}
 \label{eq:weyls-law}
\end{equation}

To specify predictive alignment, normalize Riemannian volume to an input probability measure $\varpi$ and let $\{\phi_r\}$ be the covariance eigenbasis orthonormal in $L^2(\varpi)$. For a fixed square-integrable target $f^\star=\sum_r\TargetCoefficient\phi_r$, its ideal spectral truncation $f_R^\star$ has squared error
\begin{equation}
 \|f^\star-f_R^\star\|_{L^2(\varpi)}^2
 =\sum_{r>R}|\TargetCoefficient|^2.
 \label{eq:target-spectral-tail}
\end{equation}
Predictive alignment specifies the target's power in the covariance modes. For some amplitude $b>0$, assume its coefficient power has the covariance tail profile~\cite{canatar2021alignment}. Together with \cref{eq:weyls-law}, this gives
\[
 |\TargetCoefficient|^2\sim b\lambda_r
 \sim b w_{\mathcal M}^{-\CovOrder}r^{-2\CovOrder/d}
 \qquad(r\to\infty).
\]
Identifying the predictive mass as $\lambda(r)=|\TargetCoefficient|^2$ transfers the spectral exponent and amplitude $B=bw_{\mathcal M}^{-\CovOrder}$ to the unseen-mass model.

The alignment assumption fixes both the decay exponent and its leading coefficient. Weaker bounds, $c_a\lambda_r\leq|\TargetCoefficient|^2\leq C_a\lambda_r$ with $0<c_a\leq C_a<\infty$, fix only the exponent; a random-target model with $\mathbb E|\TargetCoefficient|^2=\lambda_r$ gives the corresponding prior-averaged tail. These tails measure squared function error, which enters other losses through the bounds in \cref{sec:geometric-perspective}.

Under this predictive alignment and the earlier sampling and learning assumptions, the unseen-mass model gives
\begin{equation}
 \begin{aligned}
 \SpecExponent&=\frac{2\CovOrder}{d},\\
 \DataExponent&=\frac{2\CovOrder/d-1}{\zeta},
 \qquad \CovOrder>d/2.
 \end{aligned}
 \label{eq:beta-dimension-equivalence}
\end{equation}
Thus higher dimension spreads covariance power across more modes, making the tail slower to resolve. The trace condition ensures that their total power is finite.

For a manifold Mat\'ern family with regularity $\nu$, the covariance order is $\CovOrder=\nu+d/2$~\cite{borovitskiy2020matern}, so the same alignment condition gives $\SpecExponent=1+2\nu/d$ in \cref{eq:beta-dimension-equivalence}. Here covariance order controls spectral decay; the smoothness $s$ in \cref{sec:geometric-perspective} controls a local approximation remainder.

Regularization replaces an abrupt spectral truncation by a smooth filter. With ridge parameter $\tau>0$, the penalized approximation problem and its solution are
\[
 \begin{aligned}
 f_\tau&=\operatorname*{argmin}_{f\in\operatorname{dom}(T^{-1/2})}
 \left\{\|f-f^\star\|_{L^2(\varpi)}^2
 +\tau\|T^{-1/2}f\|_{L^2(\varpi)}^2\right\}\\
 &=T(T+\tau I)^{-1}f^\star.
 \end{aligned}
\]
Expanding in the same eigenbasis gives the squared bias and effective dimension
\begin{equation}
 \begin{aligned}
 \|f^\star-f_\tau\|_{L^2(\varpi)}^2
 &=\sum_r\left(\frac{\tau}{\lambda_r+\tau}\right)^2|\TargetCoefficient|^2,\\
 d_e(\tau)&=\sum_r\frac{\lambda_r}{\lambda_r+\tau}.
 \end{aligned}
 \label{eq:ridge-spectral-filter}
\end{equation}
Modes near $\lambda_r\approx\tau$ form a gradual resolution boundary. The bias depends on target power, whereas the effective dimension counts modes weighted by the filter. Finite-sample prediction combines this bias with sampling and noise effects~\cite{caponnetto2007rates,canatar2021alignment}.

\subsection{Information Gain and Repeated Observations}
\label{subsec:entropy-bottleneck}

Gaussian observations yield a log-determinant information gain whose derivative with respect to log sample intensity equals half the effective dimension for an orthogonal design. Retain $R$ modes with prior covariance $\Lambda_R=\operatorname{diag}(\lambda_1,\ldots,\lambda_R)$. Their coefficients and noisy observations follow
\[
 \begin{aligned}
 \boldsymbol\Theta&\sim\mathcal N(0,\Lambda_R),\\
 \boldsymbol y&=\Phi_{\PaperD}\boldsymbol\Theta+\boldsymbol\varepsilon,
 \qquad\boldsymbol\varepsilon\sim\mathcal N(0,\sigma^2I),
 \end{aligned}
\]
where row $i$ of the design matrix $\Phi_{\PaperD}$ contains the basis values at input $x_i$, and the noise is independent with variance $\sigma^2>0$. Here $I$ is the identity matrix on the observation space. Gaussian conditioning gives the posterior coefficient covariance
\[
 \Sigma_{\PaperD}
 =\left(\Lambda_R^{-1}+\sigma^{-2}
 \Phi_{\PaperD}^{\mathsf T}\Phi_{\PaperD}\right)^{-1}.
\]
Let $I_R$ denote the $R$-dimensional identity matrix. Subtracting posterior differential entropy from prior entropy yields
\begin{equation}
 I(\boldsymbol\Theta;\boldsymbol y\mid\Phi_{\PaperD})
 =\frac12\log\det\!\left(I_R+\sigma^{-2}\Lambda_R^{1/2}
 \Phi_{\PaperD}^{\mathsf T}\Phi_{\PaperD}\Lambda_R^{1/2}\right).
 \label{eq:gaussian-design-information}
\end{equation}
Thus information depends on both the covariance spectrum and which directions the observation design measures.

For an orthogonal normalized design with $R\leq\PaperD$,
\[
 \Phi_{\PaperD}^{\mathsf T}\Phi_{\PaperD}=\PaperD I_R.
\]
Write $I_R(\PaperD)$ for the information under this design and $d_{e,R}$ for the effective dimension restricted to its $R$ modes. Extending $\PaperD$ continuously as observation intensity gives
\begin{equation}
 \begin{aligned}
 I_R(\PaperD)&=\frac12\sum_{r=1}^R
 \log\!\left(1+\frac{\PaperD\lambda_r}{\sigma^2}\right),\\
 \frac{\mathrm d I_R}{\mathrm d\log\PaperD}
 &=\frac12\sum_{r=1}^R\frac{\lambda_r}{\lambda_r+\sigma^2/\PaperD}
 =\frac12d_{e,R}(\sigma^2/\PaperD).
 \end{aligned}
 \label{eq:spectral-information-rate}
\end{equation}
Here $d_{e,R}$ is the finite-mode version of \cref{eq:ridge-spectral-filter}. Increasing observation intensity lowers the spectral resolution scale $\sigma^2/\PaperD$, resolving modes with smaller eigenvalues through the same smooth filter. For nonorthogonal sampled inputs, \cref{eq:gaussian-design-information} also accounts for overlap between the measured directions.

A candidate observation's information gain increases with its posterior predictive variance. Given a candidate input $x$, write $\boldsymbol\phi(x)=(\phi_1(x),\ldots,\phi_R(x))^{\mathsf T}$. The expected information gain from its noisy response equals
\[
 \frac12\log\!\left(1+
 \frac{\boldsymbol\phi(x)^{\mathsf T}\Sigma_{\PaperD}\boldsymbol\phi(x)}{\sigma^2}
 \right).
\]
Observations in uncertain directions have larger information gain, while repeated measurements in resolved directions provide diminishing returns.

Deduplication and repeated-data experiments examine how much new evidence is gained from further training tokens~\cite{lee2021deduplication,muennighoff2023dataconstrained}, while data-pruning studies show that selection can change the learning curve itself~\cite{sorscher2022pruning}. These findings motivate comparing information-based selection with simpler rules such as duplicate-cluster membership or task-specific quality scores.

A fair comparison holds total scoring, filtering and training FLOPs fixed, so any held-out improvement represents a net benefit from selection. Repeated seeds and rare-task evaluations reveal where that benefit occurs; an oversampling ablation tests whether extra tail documents add useful signal or amplify noise. This task-loss comparison determines whether reducing posterior uncertainty improves the predictions that matter in deployment.

\begin{insight}[Spectral Decay and Information Gain]
 The independent feature-coverage model gives data-loss exponent $\DataExponent=(\SpecExponent-1)/\zeta$: steeper predictive-weight decay accelerates scaling, while greater occurrence rarity slows it. For the orthogonal Gaussian design, information gain per unit of log sample intensity equals half the effective dimension at noise scale $\sigma^2/\PaperD$.
\end{insight}

%% file: content/sec4_compute_allocation.tex
\section{Compute Allocation: Training and Deployment}
\label{sec:compute-allocation}

At fixed training compute, the optimal parameter--data split balances the marginal loss reductions from both resources. Including repeated inference changes the objective: a smaller model can justify additional training through lower serving cost at matched quality. Historical allocations, separable-loss optimization and lifecycle accounting quantify these two decisions.

\subsection{Historical Training-Compute Allocations}
\label{subsec:kaplan-law}

Kaplan et al.'s fitted allocation increases parameters faster than training tokens as compute grows, under their non-embedding parameter and training conventions~\cite{kaplan2020scaling}:
\begin{equation}
 \PaperN_{\mathrm{opt}}\propto\PaperC^{\KaplanParamExponent},
 \qquad
 \PaperD_{\mathrm{opt}}\propto\PaperC^{\KaplanDataExponent}.
 \label{eq:kaplan-allocation}
\end{equation}
Because the parameter exponent exceeds the token exponent, this fit assigns a growing share of additional compute to model size. GPT-3's \GPTThreeParamsBillion{}-billion-parameter model used roughly \GPTThreeTokensBillion{} billion training tokens, or about $\GPTThreeTokensPerParam$ tokens per parameter~\cite{brown2020fewshot}. Comparisons with later prescriptions depend on their parameter accounting, training protocols and experimental scales~\cite{pearce2024reconciling}.

\subsection{Optimal Parameter--Data Allocation}
\label{subsec:chinchilla-correction}

A separable loss model has a unique positive compute-optimal allocation that balances parameter and data contributions. The Chinchilla study motivates this approximation~\cite{hoffmann2022chinchilla}; with positive amplitudes and exponents, assume
\begin{equation}
 L(\PaperN,\PaperD)=L_\infty+
 A\PaperN^{-\ParamExponent}+B\PaperD^{-\DataExponent},
 \qquad\PaperC=\TrainFactor\PaperN\PaperD.
 \label{eq:joint-loss-compute}
\end{equation}
Here $\ParamExponent$ and $\DataExponent$ describe the loss reduction from parameters and training tokens, respectively; $\SpecExponent$ instead describes the feature spectrum. At fixed $\PaperC$, increasing model size leaves fewer tokens affordable. Substituting $\PaperD=\PaperC/(\TrainFactor\PaperN)$ and differentiating with respect to $\log\PaperN$ gives
\begin{equation}
 \ParamExponent A\PaperN^{-\ParamExponent}
 =\DataExponent B\PaperD^{-\DataExponent}.
 \label{eq:chinchilla-balance}
\end{equation}
The balance equates the marginal loss reduction from each resource. At fixed $\PaperC$, excess loss is strictly convex in $\log\PaperN$, and its first derivative changes from negative to positive as $\PaperN$ runs from zero to infinity. Thus the stationary point is the unique global minimum over positive allocations, and combining it with the budget gives
\begin{equation}
 \begin{aligned}
 \PaperN_{\mathrm{opt}}(\PaperC)
 &=\left(\frac{\ParamExponent A}{\DataExponent B}\right)^{1/(\ParamExponent+\DataExponent)}
 \left(\frac{\PaperC}{\TrainFactor}\right)^{\DataExponent/(\ParamExponent+\DataExponent)},\\
 \PaperD_{\mathrm{opt}}(\PaperC)
 &=\left(\frac{\DataExponent B}{\ParamExponent A}\right)^{1/(\ParamExponent+\DataExponent)}
 \left(\frac{\PaperC}{\TrainFactor}\right)^{\ParamExponent/(\ParamExponent+\DataExponent)}.
 \end{aligned}
 \label{eq:optimal-compute-exponents}
\end{equation}

Substituting these allocations gives the exact positive envelope coefficient
\begin{equation}
 H(A,B;\ParamExponent,\DataExponent)
 =(\ParamExponent+\DataExponent)
 \left(\frac{A}{\DataExponent}\right)^{\DataExponent/(\ParamExponent+\DataExponent)}
 \left(\frac{B}{\ParamExponent}\right)^{\ParamExponent/(\ParamExponent+\DataExponent)}.
 \label{eq:optimal-frontier-prefactor}
\end{equation}
Both loss contributions have compute exponent $\OptimalComputeExponent:=\frac{\ParamExponent\DataExponent}{\ParamExponent+\DataExponent}$. The compute-optimal loss is therefore
\begin{equation}
 L_{\mathrm{opt}}(\PaperC)-L_\infty
 =H(A,B;\ParamExponent,\DataExponent)
 \left(\frac{\TrainFactor}{\PaperC}\right)^{\OptimalComputeExponent}.
 \label{eq:optimal-compute-loss}
\end{equation}
Optimizing the parameter--data split therefore turns two loss-decay exponents into the compute-loss exponent of \cref{sec:introduction}~\cite{pearce2024reconciling}. This calculation uses matching rates for the separable loss; \cref{sec:geometric-perspective} supplies upper bounds on risk, whose tightness is a separate approximation question.

\begin{figure}[hbt!]
 \centering
 \includegraphics[width=0.88\linewidth]{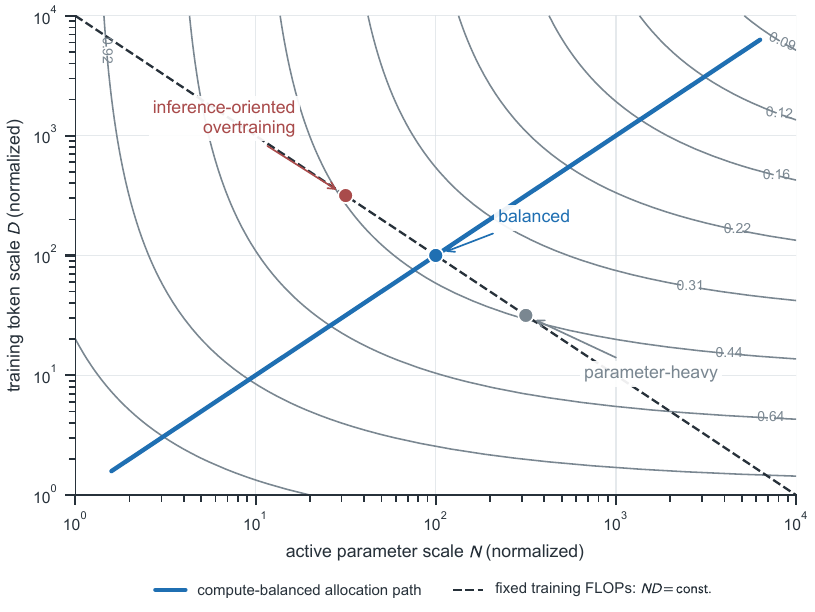}
 \caption{\textbf{Training-optimal allocation differs from serving-oriented allocation.} Gray contours hold excess loss fixed across active parameters $\PaperN$ and training tokens $\PaperD$; the balanced path (blue) meets the fixed-compute line $\PaperC=\TrainFactor\PaperN\PaperD$ (dashed, FLOP coefficient $\TrainFactor$) at minimum-loss tangency. Moving toward smaller $\PaperN$ and larger $\PaperD$ reduces serving work per token but departs from this training optimum.}
 \label{fig:compute-frontier}
\end{figure}

When the two exponents are close, optimal model size and token count grow at comparable rates. The calibration $\PaperD/\PaperN\approx\ChinchillaTokensPerParam$ is one regime-specific example of the balance in \cref{eq:chinchilla-balance}. \cref{fig:compute-frontier} shows why the balance matters: along a fixed training budget, the tangency gives the lowest loss, while allocating more tokens to a smaller model moves away from that minimum. A deployment comparison must therefore match quality before crediting the smaller model's serving savings.

Repeated epochs can improve training even when they introduce no new examples, with diminishing gains documented in repeated-corpus experiments~\cite{hernandez2022repetition,muennighoff2023dataconstrained}. Accordingly, the data coordinate in a fitted curve records either unique examples, processed tokens or epochs, so the allocation can be interpreted in the units actually varied.

\subsection{Total Compute Across Training and Deployment}
\label{subsec:pretraining-inference-blindspot}

Lifecycle compute adds a one-time training cost to a serving cost proportional to lifetime output volume~\cite{sardana2024inference}. Let $Q$ be the lifetime number of output tokens. A homogeneous dense FLOP model is
\begin{equation}
 \PaperC_{\mathrm{life}}(\PaperN,\PaperD;Q)
 =\TrainFactor\PaperN\PaperD+\InferFactor\PaperN Q.
 \label{eq:lifecycle-compute}
\end{equation}
Because training is paid once and inference recurs, a small per-token saving can repay a substantial training premium. The shared factor $\InferFactor$ assumes comparable context lengths, generation protocols and FLOP accounting across models; the measured-cost comparison below accommodates different per-token costs.

Turning FLOPs into price or latency also requires system measurements: distributed schedules change communication and utilization, while IO-aware attention changes memory traffic~\cite{narayanan2021efficient,dao2022flashattention}. Memory capacity, batching and cache behavior therefore enter deployment alongside the FLOP budget.

\subsection{Overtraining and the Deployment Crossover}
\label{subsec:overtraining-paradigm}

Additional training reduces total compute when subsequent serving savings exceed the training premium at matched quality. The Llama 3 family provides an example of small models trained beyond a local pre-training optimum~\cite{dubey2024llama3}: its rounded \LlamaSmallParamsBillion{}B parameters and \LlamaTokensTrillion{}T training tokens correspond to about $\LlamaTokensPerParam$ tokens per parameter.

Compare a small checkpoint $(\PaperN_S,\PaperD_S)$ with a large checkpoint $(\PaperN_L,\PaperD_L)$ at matched deployment quality, using one workload and resource unit. Write $\PaperC_S,\PaperC_L$ for their lifetime costs, $\PaperC_{\mathrm{train},j}$ for the training cost and $c_{\mathrm{infer},j}$ for the serving cost per output token of checkpoint $j$. The training premium and per-token serving saving give the linear cost difference
\[
 \begin{aligned}
 \TrainingCostDifference&=\PaperC_{\mathrm{train},S}-\PaperC_{\mathrm{train},L},\\
 \InferenceCostSaving&=c_{\mathrm{infer},L}-c_{\mathrm{infer},S},\\
 \PaperC_S-\PaperC_L&=\TrainingCostDifference-\InferenceCostSaving Q.
 \end{aligned}
\]
Setting this difference to zero gives the equal-cost volume for nonzero serving saving:
\begin{equation}
 \begin{aligned}
 Q^\star&=\frac{\TrainingCostDifference}{\InferenceCostSaving}
 =\frac{\PaperC_{\mathrm{train},S}-\PaperC_{\mathrm{train},L}}
 {c_{\mathrm{infer},L}-c_{\mathrm{infer},S}}\\
 &\overset{\text{dense model}}{=}
 \frac{\TrainFactor(\PaperN_S\PaperD_S-\PaperN_L\PaperD_L)}
 {\InferFactor(\PaperN_L-\PaperN_S)}.
 \end{aligned}
 \label{eq:measured-cost-crossover}
\end{equation}
The last equality specializes to the shared coefficients of \cref{eq:lifecycle-compute}; measured costs govern the general comparison. The usual overtraining trade-off has a higher upfront training cost and lower per-token serving cost, so $\TrainingCostDifference>0$ and $\InferenceCostSaving>0$. Serving savings then amortize the training premium once output volume exceeds $Q^\star$. (A smaller checkpoint with lower training cost but higher serving cost is preferable only below $Q^\star$. A checkpoint no costlier in either component and strictly cheaper in at least one dominates at every positive volume; with equal per-token costs, training cost alone decides.)

If serving demand is uncertain, the same cost difference gives both an expected saving and a break-even probability. Let $Q\geq0$ be random with finite mean, holding the matched-quality training and per-token costs fixed. Linearity gives
\begin{equation}
 \mathbb E[\PaperC_S-\PaperC_L]
 =\TrainingCostDifference-\InferenceCostSaving\,\mathbb E[Q].
 \label{eq:expected-lifecycle-difference}
\end{equation}
A risk-neutral comparison uses this expectation, so mean demand suffices when unit costs are fixed. If unit costs vary with the workload, their joint variation with demand enters the expectation as well.

In the positive-premium, positive-saving case, the probability of a saving is $\Pr(Q>Q^\star)$, which depends on the demand distribution rather than its mean alone. Expected cost and the probability of a saving therefore answer different deployment questions.

\begin{insight}[Training Balance and Deployment Crossover]
 In the separable model, optimal excess loss has compute exponent $\OptimalComputeExponent=\ParamExponent\DataExponent/(\ParamExponent+\DataExponent)$. At matched quality, a positive training premium $\TrainingCostDifference$ is repaid above output volume $Q^\star=\TrainingCostDifference/\InferenceCostSaving$ when the per-token serving saving $\InferenceCostSaving$ is positive.
\end{insight}

%% file: content/sec5_emergence_grokking.tex
\section{Emergence, Capability Thresholds and Grokking}
\label{sec:emergence-grokking}

Nonlinear scoring can produce apparent capability thresholds during smooth predictive improvement, while changes in the learned solution involve optimization dynamics. Reports of emergent abilities motivate examining both mechanisms~\cite{wei2022emergent}. Continuous scores characterize the metric effect; learning trajectories and mechanism probes characterize changes in the computation learned by the model.

\subsection{Nonlinear Scoring and Apparent Thresholds}
\label{subsec:emergent-abilities}

\begin{figure}[!htbp]
 \centering
 \includegraphics[width=\linewidth]{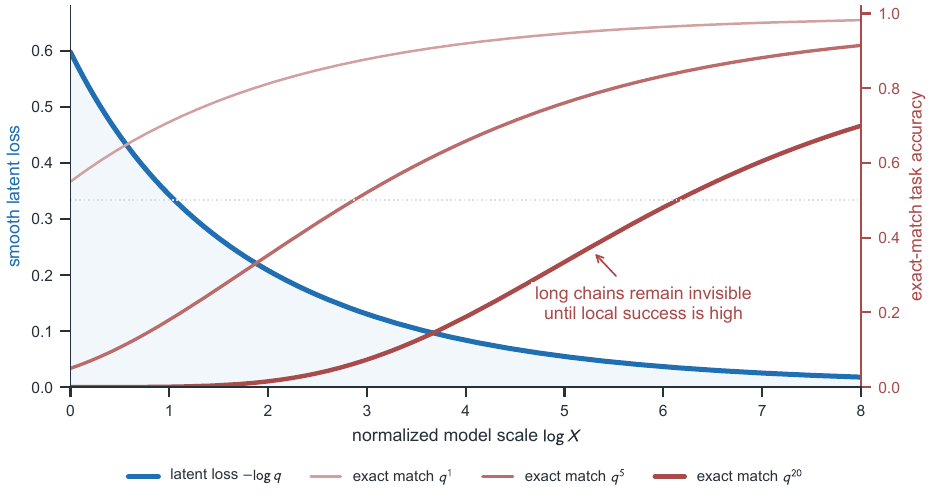}
 \caption{\textbf{Nonlinear scoring can make smooth progress appear abrupt.} As model scale $\log X$ grows, per-step success $q(X)$ improves smoothly and latent loss $-\log q$ (blue) falls. Exact-match success $q^r$ for an $r$-step chain (red) stays low until per-step reliability is high, producing sharper apparent thresholds for longer chains.}
 \label{fig:emergence-thresholds}
\end{figure}

Exact-match scoring amplifies smooth improvements in per-step reliability into sharp apparent thresholds, as metric-choice experiments demonstrate~\cite{schaeffer2023mirage}. Consider an $r$-step task whose conditional success probability at each step, given earlier success, is $q$. The chain rule gives
\begin{equation}
 P_{\mathrm{task}}(q)=q^r,\qquad
 \frac{\mathrm dP_{\mathrm{task}}}{\mathrm dq}=rq^{r-1}.
 \label{eq:multistep-threshold}
\end{equation}
For a long chain, moderate per-step success leaves exact-match accuracy near zero. A continuous change in $q$ can then appear abrupt under thresholded reporting, without a discontinuity in the underlying prediction process.
\cref{fig:emergence-thresholds} uses the same prescribed $q(X)$ for all chain lengths; the exact-match curves differ because success is scored as $q^r$.

Continuous scores and partial credit reveal progress that exact-match scoring can hide. Mechanistic progress measures and representation probes address the complementary question of when reusable solutions form~\cite{nanda2023progress,liu2022representation}. Comparing these diagnostics across model scales and seeds, with calibration and prompts controlled, separates a scoring threshold from a change in the learned computation.

\FloatBarrier
\subsection{Grokking: Training Fit Before Generalization}
\label{subsec:grokking}

Grokking is a delay between fitting the training set and generalizing to new examples~\cite{power2022grokking}. Training loss is already near zero at fitting time $t_{\mathrm{fit}}$, yet test loss improves substantially only at a later time $t_{\mathrm{gen}}\gg t_{\mathrm{fit}}$. Since networks can also fit arbitrary labels~\cite{zhang2016generalization}, a good training fit by itself leaves open which solution has been learned. Continued optimization can change that solution, with grokking observed beyond the original algorithmic tasks and influenced by initialization, regularization and task structure~\cite{liu2022omnigrok}. The central variable here is training time along one model's trajectory; model-size scaling compares different trajectories.

\subsection{Regularization Preference and Optimization Dynamics}
\label{subsec:weight-regularization}

An explicit $L_2$ penalty favors lower-norm parameters at equal training loss, while optimization dynamics determine whether that solution is reached. For parameter vector $\boldsymbol\theta$, write
\begin{equation}
 J(\boldsymbol\theta)=L_{\mathrm{train}}(\boldsymbol\theta)
 +\lambda\|\boldsymbol\theta\|_2^2,\qquad\lambda>0.
 \label{eq:regularized-objective}
\end{equation}
This objective favors a lower-norm solution among mechanisms that fit the training set equally well, as in circuit-efficiency comparisons of memorization and generalization~\cite{varma2023circuits}. Optimization must still find that solution: gradient descent started at zero remains stationary for $(x^2-1)^2$, despite lower minima elsewhere. For adaptive optimizers, the explicit penalty also differs from decoupled weight decay.

A specific dynamical result holds for unregularized logistic regression with homogeneous linear predictors on linearly separable data. With a sufficiently small positive fixed step size, gradient descent converges in direction to the $L_2$ maximum-margin separator~\cite{soudry2017implicit}. This identifies a limiting parameter direction, rather than the emergence time of a generalizing circuit in a deep network.

\Needspace{7\baselineskip}
\begin{insight}[Scoring Thresholds and Optimization Time]
 Exact-match success $P_{\mathrm{task}}=q^r$ makes longer reasoning chains more sensitive to per-step reliability. Grokking instead concerns a temporal separation $t_{\mathrm{gen}}\gg t_{\mathrm{fit}}$ along an optimization trajectory; at equal training loss, the explicit $L_2$ objective favors the lower-norm solution.
\end{insight}

%% file: content/sec6_test_time_compute.tex
\section{Test-Time Scaling: Search and Verification}
\label{sec:test-time-compute}

Independent candidate sampling gives diminishing coverage gains, while verification determines how much of that coverage becomes delivered accuracy. Test-time scaling allocates inference compute across candidate generation, refinement and selection with the model parameters held fixed. The resulting accuracy--compute trade-off depends on both proposal success and selection fidelity.

\subsection{Generation, Verification and Coordination Costs}
\label{subsec:pretraining-system-one}

Dense autoregressive generation has approximate cost proportional to parameter count and output length. The model samples $y_t\sim p_{\boldsymbol\theta}(y_t\mid x,y_{<t})$~\cite{brown2020fewshot}, giving a response of $T_{\mathrm{out}}$ tokens the cost
\begin{equation}
 \PaperC_{\mathrm{base}}\approx\InferFactor\PaperN T_{\mathrm{out}}.
 \label{eq:base-inference}
\end{equation}
Intermediate reasoning spends more generated tokens on a response~\cite{wei2022chainofthought}. Self-consistency instead aggregates sampled answers, while Tree of Thoughts adds exploration and backtracking~\cite{wang2022selfconsistency,yao2023tree}. Thus a shared checkpoint can support workflows with very different generation and coordination costs.

For $k$ proposals with representative proposal cost $\PaperC_p$, verification cost $\PaperC_v$, and orchestration cost $\PaperC_o$, a simple accounting model is
\begin{equation}
 \PaperC_{\mathrm{test}}(k)\approx k(\PaperC_p+\PaperC_v)+\PaperC_o.
 \label{eq:test-compute-requirement}
\end{equation}
For variable proposal lengths or batched verification, measured costs replace the representative per-candidate terms. Outcome verifiers and process supervision guide which answer is selected~\cite{cobbe2021verifiers,lightman2023verify}, making selection accuracy the second component of the compute--quality trade-off.

\subsection{Candidate Coverage and Selected Accuracy}
\label{subsec:probabilistic-scaling-law}

\begin{figure}[ht!]
 \centering
 \includegraphics[width=\linewidth]{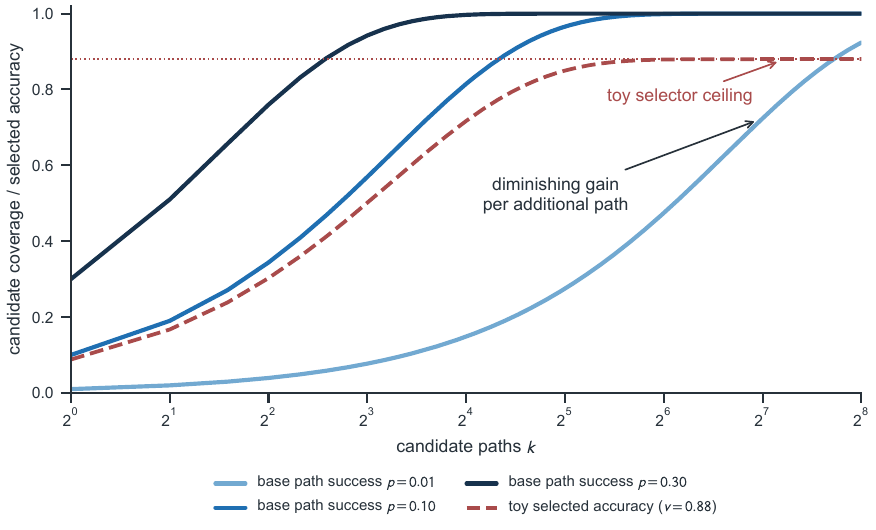}
 \caption{\textbf{Finding a correct candidate does not ensure selecting it.} For $k$ independent proposals with per-path success $p$, coverage $P_{\mathrm{cov}}(k)=1-(1-p)^k$ (solid) approaches one with diminishing gains. The red dashed example multiplies the $p=0.10$ coverage by conditional selection success $v=0.88$, giving delivered accuracy $A_{\mathrm{sel}}=vP_{\mathrm{cov}}$ and a selector-imposed ceiling.}
 \label{fig:test-time-search}
\end{figure}

Independent proposals reduce the probability that every candidate is wrong geometrically with search width~\cite{brown2024monkeys}. Fix one query $x$ and let $p$ be its single-proposal success probability. Then
\begin{equation}
 P_{\mathrm{cov}}(k)=1-(1-p)^k,\qquad
 \Delta P_{\mathrm{cov},k}
 =P_{\mathrm{cov}}(k+1)-P_{\mathrm{cov}}(k)=p(1-p)^k.
 \label{eq:test-time-success}
\end{equation}
A perfect selector converts every covered query into a correct answer, so selected accuracy equals coverage. An imperfect selector may instead choose a wrong candidate, and adding distractors can even displace a correct answer. \cref{fig:test-time-search} makes the gap visible with a constant conditional selection-success model.

For $0<p<1$ and permitted coverage failure $\delta\in(0,1)$, the minimum integer width is
\begin{equation}
 k_{\min}=\left\lceil\frac{\log\delta}{\log(1-p)}\right\rceil,
 \qquad
 k_{\min}\approx\frac{\log(1/\delta)}{p}\quad(p\ll1).
 \label{eq:test-time-required-paths}
\end{equation}
The endpoints are immediate: if $p=0$, no finite width succeeds; if $p=1$, one proposal suffices. For correlated proposals, coverage depends on their joint failure probability, which can be estimated directly or modeled through shared failure modes; pairwise correlation alone does not determine it.

Coverage must also be averaged over queries after accounting for their difficulty. If half have $p=0$ and half have $p=1$, mean coverage at $k=2$ is $1/2$, whereas substituting the mean $p=1/2$ gives $3/4$. Difficulty-stratified estimates preserve this variation before aggregation.

\subsection{Marginal Returns to Search Compute}
\label{subsec:test-time-engineering-implication}

Under independent sampling and fixed per-candidate cost, the marginal coverage gain per FLOP decreases geometrically with search width:
\begin{equation}
 \frac{\Delta P_{\mathrm{cov},k}}{\PaperC_p+\PaperC_v}
 =\frac{p(1-p)^k}{\PaperC_p+\PaperC_v}.
 \label{eq:test-time-efficiency-stopping}
\end{equation}
The geometric factor makes the marginal coverage return shrink with search width. To decide whether to continue, a serving policy uses the gain in selected-answer accuracy after the selector has acted. Inference studies find that the best compute allocation varies with query difficulty and model strength~\cite{snell2024optimal,wu2024inference}.

For selected accuracy $A_{\mathrm{sel}}(k\mid b)$ in a held-out difficulty stratum $b$, estimate the paired accuracy change and the positive incremental cost, including verification and orchestration. A practical expansion policy compares the accuracy gain per added FLOP with a minimum acceptable return $\tau_s$ set by the deployment budget or utility rule. A conservative implementation can use a calibrated lower confidence bound on the paired gain. Its confidence level controls statistical uncertainty, whereas $\tau_s$ has units of accuracy gain per FLOP and expresses the deployment tradeoff.

A held-out comparison with a fixed-budget baseline measures the policy's benefit, including the cost of repeated decisions and any latency limits. Calibration uses a reliable correctness signal for delivered answers.

\begin{insight}[Coverage Gain and Selection Return]
 Independent proposals give coverage $1-(1-p)^k$ with marginal gain $p(1-p)^k$, so additional search has diminishing returns. A deployment policy expands search when the estimated delivered-accuracy gain per added FLOP exceeds its budget-dependent threshold $\tau_s$, including generation, verification and coordination costs.
\end{insight}

%% file: content/sec7_beyond_llms.tex
\section{Extending the Framework Beyond LLMs}
\label{sec:beyond-llms}

Visual resolution raises attention cost quadratically in token count, while removing global rigid pose reduces the dimension of regular molecular configurations by six. Both effects change the balance between representation accuracy and compute. Independent conformational coverage introduces a further data constraint in molecular prediction.

\subsection{Visual Resolution: Prediction Error and Compute Cost}
\label{subsec:vlm-videomae}

At fixed depth and width, dense visual attention cost grows quadratically with the number of image patches or video tubelets, as used in Vision Transformers and VideoMAE~\cite{dosovitskiy2020vit,tong2022videomae}. For image dimensions $H,W$, spatial patch width $P$, frame count $T$ and temporal tubelet width $\tau$, a divisible grid has token count
\[
 M_{\mathrm{vis}}=\frac{HW}{P^2}\frac{T}{\tau}.
\]
Squaring this sequence length gives the attention dependence:
\begin{equation}
 \PaperC_{\mathrm{attn}}\propto
 \left(\frac{HW}{P^2}\frac{T}{\tau}\right)^2.
 \label{eq:visual-token-count}
\end{equation}
Halving $P$ therefore multiplies tokens by four and dense attention work by sixteen when all other dimensions are fixed. The full training cost also includes tokenwise operations and the decoder, which we account for below.

Vision scaling studies jointly vary model, data and compute~\cite{zhai2021vision}. Within a fixed spatial/temporal sampling protocol, let $R=M_{\mathrm{vis}}$ index token resolution and let $\PaperD$ count example presentations. A local separable loss ansatz is
\begin{equation}
 L(\PaperN,\PaperD,R)=L_\infty+
 A_\PaperN\PaperN^{-\ParamExponent}+A_\PaperD\PaperD^{-\DataExponent}
 +A_RR^{-\alpha_R}.
 \label{eq:multimodal-scaling-surface}
\end{equation}
Holding aspect ratio and temporal sampling fixed gives token count a consistent meaning as resolution; changing those protocols changes which details the tokens contain.

A fixed-depth dense Transformer with negligible embedding contribution has parameter count proportional to hidden width squared. Tokenwise work scales with parameter count, while the quadratic attention term scales linearly with width; replacing width by $\sqrt{\PaperN}$ gives the following local cost model:
\begin{equation}
 \PaperC_{\mathrm{vis}}(\PaperN,\PaperD,R)
 =\PaperD\left(a\PaperN R+b\sqrt{\PaperN}R^2\right)+\PaperC_{\mathrm{dec}},
 \quad a,b>0.
 \label{eq:visual-cost-surface}
\end{equation}
The terms represent tokenwise work, quadratic attention and a separately accounted decoder.

The optimal budget split across $\PaperN$, $\PaperD$ and $R$ equalizes their marginal loss reduction per unit of full compute. Differentiating the cost includes decoder derivatives whenever decoder work changes with an allocation; a fixed decoder workload contributes a constant instead. At a differentiable interior optimum with positive cost derivatives, let $\lambda_{\mathrm{cost}}$ denote the common marginal return, or Lagrange multiplier:
\begin{equation}
 \frac{-\partial_xL}{\partial_xC_{\mathrm{vis}}}=\lambda_{\mathrm{cost}},
 \qquad x\in\{\PaperN,\PaperD,R\}.
 \label{eq:visual-interior-allocation}
\end{equation}
This condition identifies an interior allocation. At boundaries or discrete patch choices, feasible comparisons or KKT conditions determine the optimum using a fitted loss surface and a profiled workload.

\Needspace{7\baselineskip}
Masked autoencoders reduce the encoder input while retaining a reconstruction decoder~\cite{he2021masked,tong2022videomae}. For mask fraction $\mu\in[0,1)$,
\begin{equation}
 M_{\mathrm{enc}}=(1-\mu)M_{\mathrm{vis}},\qquad
 \frac{\PaperC_{\mathrm{attn,enc}}}{\PaperC_{\mathrm{attn}}}\approx(1-\mu)^2.
 \label{eq:masked-video-compute}
\end{equation}
The encoder attention saving follows from shortening both axes of its attention matrix; decoder and tokenwise work remain in the full cost.

A paired held-out comparison at fixed $\PaperN,\PaperD$ and sampling protocol measures how much extra resolution improves the task. Repeated seeds quantify uncertainty in that gain, while the full workload, including the decoder, gives its computational price.

\subsection{Molecular Symmetry, Compute and Conformational Coverage}
\label{subsec:molecular-entropy-bottleneck}

Rigid-motion symmetry reduces the representation dimension in molecular prediction, while structural observations determine which conformations can be learned. Protein structure models infer atomic arrangements from sequence and biological context~\cite{jumper2021alphafold}; the AlphaFold stages in \cref{tab:alphafold-stages} expand the target scope and change the coordinate construction. AF2 uses local frames and invariant point attention, whereas AF3's diffusion module operates directly on atom coordinates without rotational frames or equivariant processing~\cite{abramson2024alphafold3}. The following calculations quantify the approximation gain from removing rigid motion and its trade-off with architecture cost and conformational coverage.

\begin{table}[htbp]
 \centering\small
 \caption{AlphaFold stages expand the prediction target and change the mechanism used to construct structures.}
 \label{tab:alphafold-stages}
 \begin{tabular}{@{}p{0.12\linewidth}p{0.29\linewidth}p{0.51\linewidth}@{}}
 \toprule Stage & Evidence/target & Output construction\\ \midrule
 AF1~\cite{senior2020alphafold} & Sequence-derived evidence; protein structure & Predicted distance distributions guide structure optimization\\
 AF2~\cite{jumper2021alphafold} & Alignments, pair representations; protein structure & Evoformer with an end-to-end structure module\\
 AF3~\cite{abramson2024alphafold3} & Joint biomolecular complexes & Diffusion-based coordinate generation\\
 \bottomrule
 \end{tabular}
\end{table}

Rotating or translating a molecule changes its pose while preserving its internal shape. For labelled, non-collinear configurations on a regular $d$-dimensional configuration space $\mathcal M$, the free, proper action of $G=\mathrm{SE}(3)$ has orbit dimension $\RigidDim=6$. Removing the center of mass and then quotienting rotations gives the shape space $\mathcal B=\mathcal M/G$, with dimension
\[
 \InternalDim=\dim(\mathcal B)
 =\dim(\mathcal M)-\dim G=d-6>0.
\]
A compact shape domain can then be selected. An invariant scalar target, such as internal energy, satisfies $f^\star(gx)=f^\star(x)$, so it can be represented on shape space alone. For equivariant coordinate outputs, local-frame components represent shape and reconstruct the pose when producing coordinates.

The covering model of \cref{sec:geometric-perspective} quantifies the cost of resolving these coordinates. Compare a regular local construction covering bounded pose charts with one resolving internal shape directly, with common smoothness $s$ and capacity exponent $\gamma$. For $j\in\{\mathrm{pose},\mathrm{shape}\}$ set $d_{\mathrm{pose}}:=d$ and $d_{\mathrm{shape}}:=\InternalDim$. Let $C_j,c_s^{(j)}$ and $\PartitionFactor^{(j)}$ be the corresponding covering, remainder and capacity constants. Using the same integer-budget envelope as \cref{subsec:parameters-resolution}, set $M_j:=c_s^{(j)}C_j^s(\PartitionFactor^{(j)})^{-s/d_j}$. \cref{eq:parameter-error-bound} then gives
\begin{equation}
 \begin{aligned}
 \ApproxError_{\mathrm{pose}}&\leq M_{\mathrm{pose}}\PaperN_e^{-\gamma s/d},\\
 \ApproxError_{\mathrm{shape}}&\leq M_{\mathrm{shape}}\PaperN_e^{-\gamma s/\InternalDim}.
 \end{aligned}
 \label{eq:quotient-error-bounds}
\end{equation}
The relative increase in the exponents of these envelopes is
\begin{equation}
 \frac{\gamma s/\InternalDim}{\gamma s/d}-1
 =\frac{\RigidDim}{\InternalDim}.
 \label{eq:quotient-approximation-gain}
\end{equation}
At fixed $s,\gamma,\RigidDim$, the absolute exponent difference obeys
\[
 \frac{\gamma s}{d-\RigidDim}-\frac{\gamma s}{d}
 =\frac{\gamma s\RigidDim}{d(d-\RigidDim)}
 \sim\frac{\gamma s\RigidDim}{d^2}
 \qquad(d\to\infty).
\]
Thus the relative gain $6/\InternalDim$ shrinks as internal complexity grows: removing six pose coordinates changes a low-dimensional problem much more than a high-dimensional one. Squared-loss risk bounds inherit twice these exponents, and a learner that discovers the symmetry can also exploit the reduced representation.

Shape space also supplies the spectral modes for an invariant target. Suppose $\mathcal B$ is a smooth compact Riemannian manifold without boundary, and use its nonnegative self-adjoint Laplace--Beltrami operator $\Delta_{\mathcal B}$. For $a>0$ and $\CovOrder>\InternalDim/2$, take $T_{\mathcal B}=(aI+\Delta_{\mathcal B})^{-\CovOrder}$. With $w_{\mathcal B}$ the Weyl constant of \cref{eq:weyls-law}, replacing $\mathcal M,d$ by $\mathcal B,\InternalDim$ gives
\begin{equation}
 \begin{aligned}
 \mu_r^{\mathcal B}&\sim w_{\mathcal B}r^{2/\InternalDim},\\
 \lambda_r^{\mathcal B}&\sim w_{\mathcal B}^{-\CovOrder}r^{-2\CovOrder/\InternalDim},\\
 \DataExponent^{\mathcal B}&=\frac{2\CovOrder/\InternalDim-1}{\zeta}.
 \end{aligned}
 \label{eq:quotient-spectral-rate}
\end{equation}
The data exponent uses predictive alignment in these modes and the observation model of \cref{sec:information-theory}. Because an invariant target has zero coefficients in symmetry-forbidden directions, its predictive spectrum is ordered within shape space. The regular-orbit assumption fixes the quotient dimension; collinear configurations have rotational stabilizers and form a different stratum.

\begin{figure}[ht!]
 \centering
 \includegraphics[width=\linewidth]{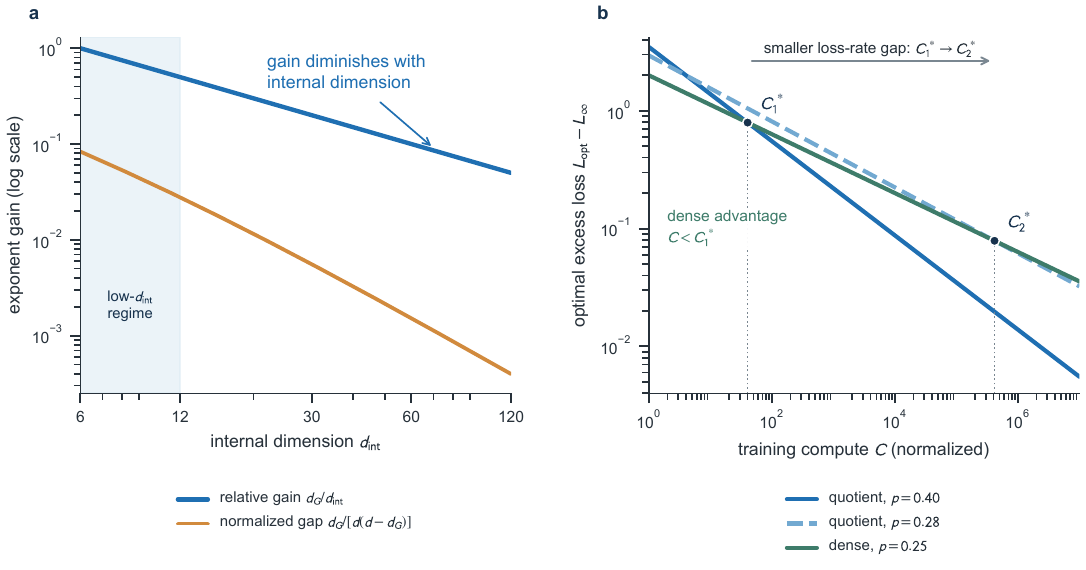}
 \caption{\textbf{Dimension dilutes geometric gains; compute cost shifts crossovers.} \textbf{(a)} Removing $\RigidDim=6$ rigid-pose coordinates from total dimension $d$ leaves internal dimension $\InternalDim=d-\RigidDim$; both relative and normalized exponent gains decline. \textbf{(b)} Optimal excess loss scales as $L_{\mathrm{opt}}-L_\infty\propto\PaperC^{-p}$: the costlier quotient model wins above each crossing $C_i^\star$, while a smaller loss-exponent gap delays that crossing.}
 \label{fig:symmetry-compute}
\end{figure}

To compare architecture costs, apply the separable loss model at one target, metric and floor. For $j\in\{\mathrm{eq},\mathrm{dense}\}$, let $A_j,B_j>0$ be loss amplitudes, $\ParamExponent^{(j)},\DataExponent^{(j)}>0$ the task-loss exponents, and $\TrainFactor^{(j)}$ the FLOP coefficient in $\PaperC=\TrainFactor^{(j)}\PaperN\PaperD$. The allocation result gives the compute-loss exponent
\[
 p_j=\frac{\ParamExponent^{(j)}\DataExponent^{(j)}}
 {\ParamExponent^{(j)}+\DataExponent^{(j)}}.
\]
Using the closed coefficient $H_j:=H(A_j,B_j;\ParamExponent^{(j)},\DataExponent^{(j)})$ from \cref{eq:optimal-frontier-prefactor}, the optimal frontier is
\begin{equation}
 L_{\mathrm{opt},j}(\PaperC)-L_\infty
 =H_j\left(\frac{\TrainFactor^{(j)}}{\PaperC}\right)^{p_j}.
 \label{eq:architecture-compute-frontier}
\end{equation}
With these task-loss exponents, dividing the two frontiers directly compares the error achieved at equal compute:
\begin{equation}
 \begin{aligned}
 \frac{L_{\mathrm{opt},\mathrm{dense}}(\PaperC)-L_\infty}
 {L_{\mathrm{opt},\mathrm{eq}}(\PaperC)-L_\infty}
 &=\frac{H_{\mathrm{dense}}}{H_{\mathrm{eq}}}
 \frac{(\TrainFactor^{(\mathrm{dense})})^{p_{\mathrm{dense}}}}
 {(\TrainFactor^{(\mathrm{eq})})^{p_{\mathrm{eq}}}}
 \PaperC^{p_{\mathrm{eq}}-p_{\mathrm{dense}}}<1.
 \end{aligned}
 \label{eq:architecture-compute-comparison}
\end{equation}
The dense model has lower loss when this ratio is below one. \cref{fig:symmetry-compute}(a) shows how approximation-exponent gains decline with dimension; panel (b) compares compute frontiers at two loss-exponent gaps, using independently set loss exponents and cost coefficients.

\FloatBarrier

A fixed compute premium raises a frontier's intercept, while a better loss exponent steepens its descent. Equal loss exponents give a budget-independent comparison; with unequal exponents, the steeper frontier eventually wins. Applying this trade-off to AF2- and AF3-style constructions requires their task-loss amplitudes, exponents and FLOP coefficients on matched tasks. Hardware throughput enters the corresponding time-cost comparison.

Conformational coverage depends on independent structural evidence as well as representation accuracy. The Protein Data Bank (PDB) archives experimental structures~\cite{berman2000pdb}, whereas AlphaFold DB supplies predicted structures~\cite{varadi2024database}. Repeated records can sharpen estimates for a known state; related families and model predictions provide shared or inferred information. Independent conformational coverage depends on the states actually observed, and rare states can remain poorly covered as a database grows.

\begin{figure}[ht!]
 \centering
 \includegraphics[width=\linewidth]{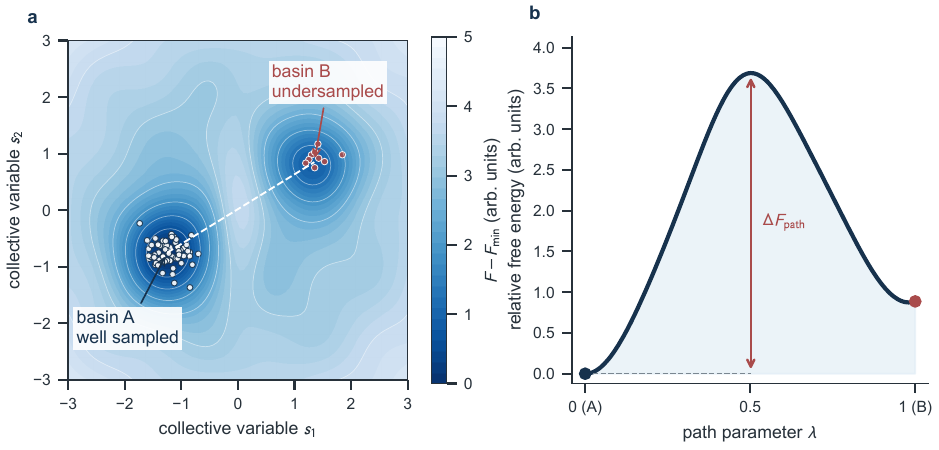}
 \caption{\textbf{Rare conformations can remain poorly covered.} \textbf{(a)} Example observations occupy basin A more densely than B on a free-energy surface in collective variables $s_1,s_2$. \textbf{(b)} Along path parameter $\lambda$, $\Delta F_{\mathrm{path}}$ marks the maximum above A; equilibrium rarity depends on integrated basin weights. More records need not resolve rare basins.}
 \label{fig:molecular-entropy-bottleneck}
\end{figure}

For independent equilibrium draws, basin free energies determine how often each conformation appears. AF3's multi-seed diffusion protocol instead generates and ranks structural candidates, rather than sampling equilibrium basin weights~\cite{abramson2024alphafold3}. At fixed context and temperature $T_{\mathrm{phys}}$, let $p_{\mathrm{dom}}$ and $p_{\mathrm{alt}}$ be the dominant and alternative basin probabilities. Their integrated free-energy difference $\Delta F>0$ gives the Boltzmann ratio, with Boltzmann constant $k_B$:
\[
 \frac{p_{\mathrm{alt}}}{p_{\mathrm{dom}}}
 =\exp\!\left(-\frac{\Delta F}{k_BT_{\mathrm{phys}}}\right).
\]
The minimum integer sample count giving an expected count of at least one in the alternative basin is
\begin{equation}
 \PaperD_{\mathrm{observe}}=\left\lceil\frac{1}{p_{\mathrm{alt}}}\right\rceil
 =\left\lceil\frac{e^{\Delta F/(k_BT_{\mathrm{phys}})}}{p_{\mathrm{dom}}}\right\rceil.
 \label{eq:rare-conformation-sample-requirement}
\end{equation}
The required count grows exponentially with the basin free-energy gap and gives an order-one observation chance under independent equilibrium draws. Correlation changes free-energy estimation~\cite{shirts2008freeenergy}, while experimental structure databases follow their own collection protocols. \cref{fig:molecular-entropy-bottleneck}(a) visualizes uneven state coverage; panel (b) marks a path maximum above the starting basin, whereas \cref{eq:rare-conformation-sample-requirement} uses the integrated free energies of the basins themselves.

Repeated measurements can improve precision even when they add no new states. To quantify this effect for one common scalar estimand, suppose cluster $j$ contains $n_j$ measurements, clusters are independent, individual variances equal $\sigma^2$, and within-cluster correlations are exchangeable $\ClusterCorr\in[0,1]$. Summing individual variances and pairwise covariances gives the variance of the cluster mean $\bar Y_j$:
\[
 \operatorname{Var}(\bar Y_j)
 =\frac{\sigma^2}{n_j^2}
 \left[n_j+n_j(n_j-1)\ClusterCorr\right]
 =\frac{\sigma^2[1+(n_j-1)\ClusterCorr]}{n_j}.
\]
Independent cluster precisions add under inverse-variance weighting; dividing the resulting precision by $1/\sigma^2$ gives equivalent precision count
\begin{equation}
 \PrecisionCount=\sum_j\frac{n_j}{1+(n_j-1)\ClusterCorr}.
 \label{eq:effective-structural-data}
\end{equation}

The precision count $\PrecisionCount$ measures the accuracy of this inverse-variance-weighted estimator. Equal record weighting generally gives a different effective sample size, while counting covered structural states answers a different question. Precision, unique targets and state coverage therefore describe complementary aspects of a structural dataset.

\FloatBarrier
\subsection{Normalized Loss and Cross-Architecture Comparisons}
\label{subsec:non-llm-methodology}

Normalizing excess loss at a reference scale makes fractional improvements comparable across metrics with different units. Next-token cross-entropy, reconstruction error and structural error have different floors, and AF2's frame-aligned point error also depends on alignment, normalization and clipping~\cite{jumper2021alphafold}. Let $L_j(X)$ be metric $j$'s loss at resource scale $X$, with floor $L_{j,\infty}$. Within a fixed run family, a reference scale $X_{\mathrm{ref}}$ with positive excess loss gives
\begin{equation}
 \widetilde\epsilon_j(X)=
 \frac{L_j(X)-L_{j,\infty}}{L_j(X_{\mathrm{ref}})-L_{j,\infty}}.
 \label{eq:molecular-normalized-excess-loss}
\end{equation}
If the excess loss follows $A_jX^{-\alpha_j}$ on the comparison interval, this normalization yields $\widetilde\epsilon_j(X)=(X/X_{\mathrm{ref}})^{-\alpha_j}$. In general it compares fractional reductions within each metric, while the scientific meaning remains task-specific.

Cross-architecture comparisons then hold data splits, related-family controls, training cutoffs, masks and clipping conventions fixed~\cite{jumper2021alphafold,abramson2024alphafold3}. Active and resident parameters, processed training examples or tokens, unique evaluation targets and measured FLOPs describe the workload together, because dense Transformers, state-space models and sparse mixtures execute different operations per token.

Inference adds two computational controls: denoising steps refine one candidate, while extra samples expand the set available for selection (\cref{sec:test-time-compute}). Denoising iterations index the generation algorithm rather than physical evolution. In AF3's low-homology antibody--antigen evaluation, interfaces are ranked by protein--protein ipTM and success is measured with DockQ; extra seeds produce a pronounced benefit in this task, with different effects across other molecule classes~\cite{abramson2024alphafold3}. Held-out structural correctness measures the value of confidence-based selection, while equilibrium basin coverage requires a physical sampling model.

\subsection{Representation, Data and Resource Allocation}
\label{subsec:conclusions}

Scaling rates depend jointly on representation resolution, feature coverage and the compute allocated to reducing each source of error. Finer partitions reduce function error according to smoothness and dimension; additional observations reduce the residual from rare predictive features. Compute constraints balance these loss reductions, while scoring and selection determine the reported accuracy. Visual resolution adds attention work, and molecular prediction couples symmetry-reduced representation with independent conformational coverage.

These mechanisms explain why a scaling exponent depends on the target, loss, data and architecture together. Held-out resource scales test whether the fitted rate predicts further improvement; matched-quality comparisons determine whether that improvement repays its training and serving cost. Moving to a new domain begins with its prediction target and sources of independent evidence, from which the appropriate resource curve can be derived or measured.